%% file: project.tex
\documentclass[sigconf]{acmart}
\usepackage{graphicx} % insert figure
\usepackage{lipsum}  % generate context
\usepackage{xspace}
\usepackage{xcolor}
\usepackage{tabularx}
\usepackage{booktabs}
\usepackage{placeins}
\usepackage{amsfonts}

\usepackage{algorithmic}
\usepackage{textcomp}
\usepackage{amsmath,bm}
\usepackage{pbox}
\usepackage{amssymb}
\usepackage{makecell}
\usepackage{lipsum}
\usepackage{caption, subcaption}
\usepackage{textcomp}
\usepackage{upgreek}
\usepackage{algorithm}
\usepackage{url}
\usepackage{multirow}
\usepackage{enumitem,soul}
\usepackage{mathtools,lipsum}
\usepackage{cuted}
\usepackage{booktabs}
\usepackage{subcaption}

\usepackage{algorithm}
\makeatletter
\DeclareRobustCommand\onedot{\futurelet\@let@token\@onedot}
\def\@onedot{\ifx\@let@token.\else.\null\fi\xspace}

\def\eg{\emph{e.g}\onedot} 
\def\ie{\emph{i.e}\onedot}

\makeatother

\AtBeginDocument{%
  \providecommand\BibTeX{{%
    \normalfont B\kern-0.5em{\scshape i\kern-0.25em b}\kern-0.8em\TeX}}}

\setcopyright{acmlicensed}
\copyrightyear{2018}
\acmYear{2018}
\acmDOI{XXXXXXX.XXXXXXX}

\acmConference[KDD '25]{KDD '25: ACM SIGKDD Conference on Knowledge Discovery and Data Mining}{August 3-7, 2025}{Toronto, Canada}
\acmISBN{978-1-4503-XXXX-X/18/06}

\begin{document}

% \title{Conditional Spatial-Temporal Contrastive Learning for Generic Spatial-Temporal Representations}
\title{
UrbanMind: Urban Dynamics Prediction with Multifaceted Spatial-Temporal Large Language Models}

% \author{Anonymous Authors}
% \author{
% Yuhang Liu\textsuperscript{1},
% Yingxue Zhang\textsuperscript{1},
% Xin Zhang\textsuperscript{2},
% Ling Tian\textsuperscript{3},
% Xu Zheng\textsuperscript{3},
% Yanhua Li\textsuperscript{4},
% Jun Luo\textsuperscript{5} \\
% \textsuperscript{1}Binghamton University,
% \textsuperscript{2}San Diego State University,
% \textsuperscript{3}University of Electronic Science and Technology of China,
% \textsuperscript{4}Worcester Polytechnic Institute,\\
% \textsuperscript{5}Logistics and Supply Chain MultiTech R\&D Centre \\
% \texttt{\{yliu41,yzhang42\}@binghamton.edu}, \texttt{xzhang19@sdsu.edu}, \texttt{\{lingtian,xzheng\}@uestc.edu.cn},
% \texttt{yli15@wpi.edu}, \texttt{jluo@lscm.hk}
% }

\author{Yuhang Liu}
\email{yliu41@binghamton.edu}
\affiliation{
  \institution{State University of New York at Binghamton}
  \city{Binghamton}
  \state{NY}
  \country{USA}
}

\author{Yingxue Zhang}
\authornote{Corresponding author.}
\email{yzhang42@binghamton.edu}
\affiliation{
  \institution{State University of New York at Binghamton}
  \city{Binghamton}
  \state{NY}
  \country{USA}
}

\author{Xin Zhang}
\email{xzhang19@sdsu.edu}
\affiliation{
  \institution{San Diego State University}
  \city{San Diego}
  \state{CA}
  \country{USA}
}

\author{Ling Tian}
\email{lingtian@uestc.edu.cn}
\affiliation{
  \institution{University of Electronic Science and Technology of China}
  \city{Chengdu}
  \country{China}
}

% \author{Xu Zheng}
% \email{xzheng@uestc.edu.cn}
% \affiliation{
%   \institution{University of Electronic Science and Technology of China}
%   \city{Chengdu}
%   \country{China}
% }

\author{Yanhua Li}
\email{yli15@wpi.edu}
\affiliation{
  \institution{Worcester Polytechnic Institute}
  \city{Worcester}
  \state{MA}
  \country{USA}
}

\author{Jun Luo}
\email{jluo@lscm.hk}
\affiliation{
  \institution{Logistics and Supply Chain MultiTech R\&D Centre}
  \city{Hong Kong}
  \country{China}
}

\renewcommand{\shortauthors}{Anonymous.}


\input{Content/abstract}






%%
%% This command processes the author affiliation and title
\maketitle

% === 自定义 Reference Format（显示在正文最后）===
% \section*{ACM Reference Format}
% \small
% Yuhang Liu\textsuperscript{1}, Yingxue Zhang\textsuperscript{1}, Xin Zhang\textsuperscript{2}, Ling Tian\textsuperscript{3}, Xu Zheng\textsuperscript{3}, Yanhua Li\textsuperscript{4}, and Jun Luo\textsuperscript{5}.  
% 2025. *UrbanMind: Urban Dynamics Prediction with Multifaceted Spatial-Temporal Large Language Models*.  
% In *Proceedings of ACM (KDD '25)*, August 3–7, 2025, Toronto, Canada. ACM, 11 pages.  
% https://doi.org/XXXXXXX

\begin{figure}[ht]
\includegraphics[width=0.45\textwidth]{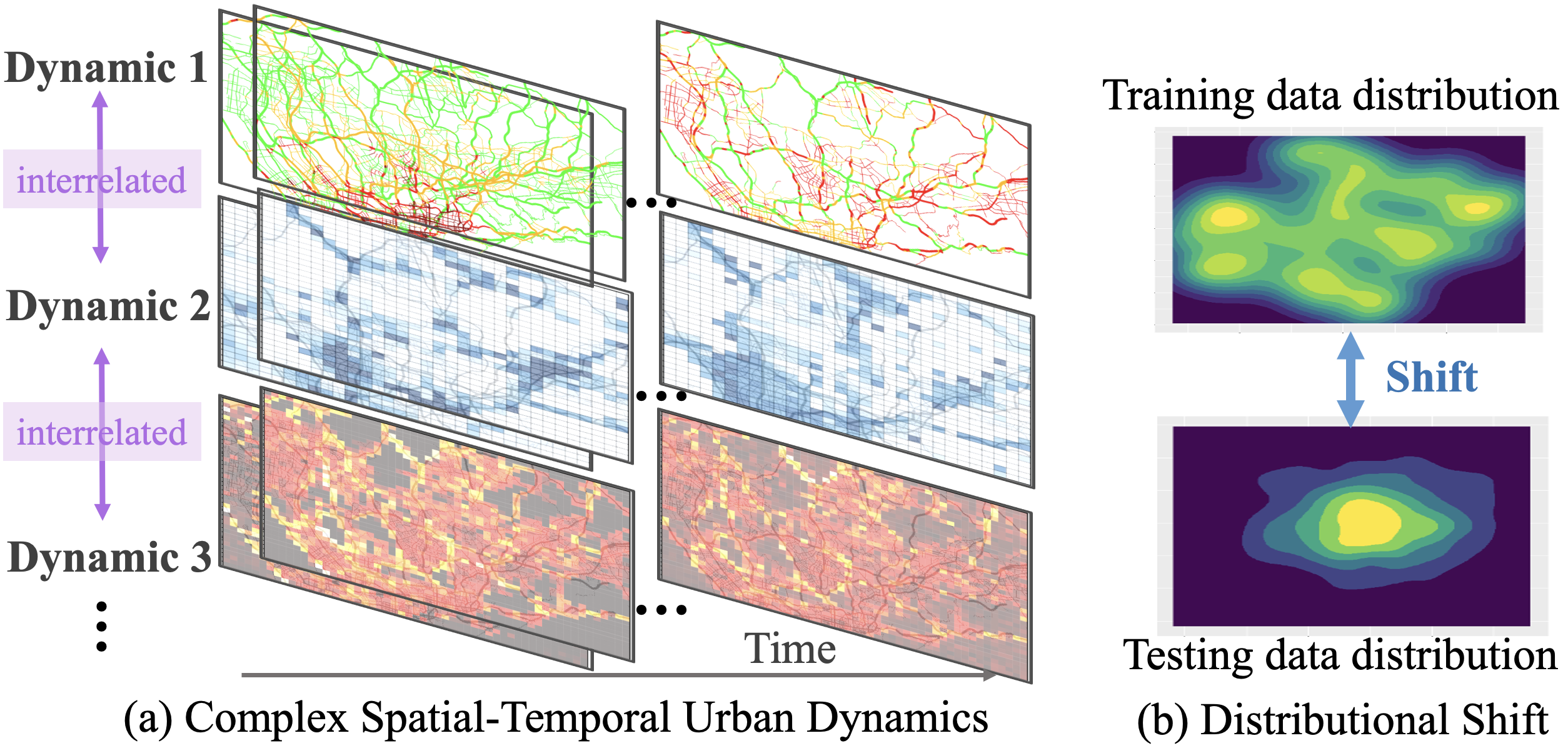}
\vspace{-0.35cm}
\caption{Illustration of key challenges, including complex urban dynamics and distributional shifts.
}
\label{fig:intro}
\vspace{-20pt}
\end{figure}

\input{Content/introduction}

\input{Content/overview}

\input{Content/method}

\input{Content/experiment}

\input{Content/related}

\section{Conclusion}
In this paper, we present UrbanMind, a novel spatial-temporal LLM designed to effectively process and understand the intricate relationships and patterns within spatial-temporal data for urban dynamics prediction. UrbanMind demonstrates high prediction accuracy and robust generalization, including in zero-shot scenarios where no prior data is available. Its key innovations include a masked-empowered representation learning framework implemented through the novel Muffin-MAE, which employs advanced masking strategies to capture complex spatial-temporal dependencies and inter-correlations across multi-faceted urban dynamics. Additionally, UrbanMind incorporates a semantic-rich prompt design and fine-tuning strategy tailored for spatial-temporal data, as well as a testing-time adaptation mechanism that mitigates distributional shifts through a reconstruction module.
Extensive experiments on three urban dynamics—traffic speed, inflow, and travel demand—across three cities validate UrbanMind's effectiveness. The results highlight its superior generalization and predictive accuracy across diverse spatial-temporal scenarios, consistently outperforming state-of-the-art baselines, even in unseen regions or data-scarce settings.

\section{Acknowledgments}
\noindent Yanhua Li was supported in part by NSF grants IIS-1942680 (CAREER), CNS-1952085 and DGE-2021871. Jun Luo is supported by The Innovation and Technology Fund (Ref. ITP/012/25LP).

\noindent \textbf{Disclaimer:}
Any opinions, findings, conclusions or recommendations expressed in this material/event (or by members of the project team) do not reflect the views of the Government of the Hong Kong Special Administrative Region, the Innovation and Technology Commission or the Innovation and Technology Fund Research Projects Assessment Panel.

\newpage

\bibliographystyle{ACM-Reference-Format}
\bibliography{project,reference}

%\newpage
\appendix
\input{Content/appendix}

\end{document}

%% file: Content/abstract.tex
\begin{abstract}
Understanding and predicting urban dynamics is crucial for managing transportation systems, optimizing urban planning, and enhancing public services. While neural network-based approaches have achieved success, they often rely on task-specific architectures and large volumes of data, limiting their ability to generalize across diverse urban scenarios. Meanwhile, Large Language Models (LLMs) offer strong reasoning and generalization capabilities, yet their application to spatial-temporal urban dynamics remains underexplored. Existing LLM-based methods struggle to effectively integrate multifaceted spatial-temporal data and fail to address distributional shifts between training and testing data, limiting their predictive reliability in real-world applications.
To bridge this gap, we propose UrbanMind, a novel spatial-temporal LLM framework for multifaceted urban dynamics prediction that ensures both accurate forecasting and robust generalization. At its core, UrbanMind introduces Muffin-MAE, a multifaceted fusion masked autoencoder with specialized masking strategies that capture intricate spatial-temporal dependencies and intercorrelations among multifaceted urban dynamics. Additionally, we design a semantic-aware prompting and fine-tuning strategy that encodes spatial-temporal contextual details into prompts, enhancing LLMs' ability to reason over spatial-temporal patterns. To further improve generalization, we introduce a test time adaptation mechanism with a test data reconstructor, enabling UrbanMind to dynamically adjust to unseen test data by reconstructing LLM-generated embeddings.
Extensive experiments on real-world urban dynamics datasets from multiple cities demonstrate the effectiveness of UrbanMind. The results consistently show that UrbanMind outperforms state-of-the-art baselines, achieving superior accuracy and strong generalization, even in zero-shot scenarios with no prior data.

\end{abstract}

%% file: Content/introduction.tex
\section{Introduction}\label{sec:introduction}
%The core motivation could center around the fundamental challenge of capturing complex interdependencies in spatial-temporal data while maintaining model adaptability. 

%leverages LLMs' inherent pattern recognition and predictive capabilities while working with carefully engineered spatial-temporal embeddings. The key innovation lies in how you transform spatial-temporal data into a format that allows LLMs to effectively process and generate predictions.
Urban dynamics prediction in traffic systems involves predicting human mobility patterns, such as traffic speed and travel demand, using historical spatial-temporal data collected from various IoT devices mounted on vehicles and public transports. %It is critical for modern urban environments, enabling cities to anticipate and respond to changes in transportation patterns, population movement, and resource utilization. 
Accurate prediction holds significant potential for optimizing traffic management, enhancing urban planning, and improving public service delivery. 
%For instance, graph-based neural networks including LSGCN\cite{LSGCN}, STGCN~\cite{Yu_2018} and SAMSGL~\cite{samsgl} have been particularly successful in modeling spatial-temporal dependencies, providing stable short- and long-term predictions. Generative models have also gained traction in addressing diverse urban scenarios. TrafficGAN~\cite{ICDMtrafficGAN} and Curb-GAN~\cite{curbgan} employ adversarial learning frameworks to estimate traffic patterns under various conditions. Recent advancements, such as DYffusion~\cite{cachay2024dyffusion}, STGAIL~\cite{liu2024align} and USTD~\cite{hu2024towards}, leverage diffusion processes to model probabilistic spatial and temporal dynamics. Meta-learning-based models have additionally been explored for their flexibility and predictive power, TD$^2$-DL~\cite{TD2-DL}, cST-ML~\cite{yingxue2020cstml} and DAC-ML~\cite{zhang2021dac} adapt effectively to dynamically changing traffic conditions.

\noindent\textbf{Limitations of NN-based approaches.} Various deep neural network (NN)-based approaches have been developed for urban dynamics prediction. Graph-based neural networks \cite{LSGCN, Yu_2018, zou2024samsgl} have demonstrated exceptional success in modeling spatial-temporal dependencies. Generative models, including GAN-based methods \cite{ICDMtrafficGAN, curbgan, zhang2021c3} and diffusion-based models \cite{cachay2024dyffusion, liu2024align, hu2024towards}, have gained significant attention for their ability to model complex urban data distributions. Meta-learning and transfer-learning-based approaches \cite{TD2-DL, yingxue2020cstml, zhang2021dac, zhang2022strans, mestgan} have also been explored for their flexibility and adaptability. Despite these advances, existing methods often rely heavily on task-specific architectures and substantial volumes of data. These dependencies limit their ability to generalize effectively across diverse urban contexts or adapt to evolving conditions, particularly in scenarios involving unseen regions or sparse data.
This limitation raises the question: \textit{How can we enhance urban dynamics prediction by simultaneously achieving high accuracy and strong generalization across dynamic and diverse urban scenarios?}

\noindent\textbf{Limitations of LLM-based approaches.} Considering the outstanding generalization ability, we turn attention to Large Language Models (LLMs)~\cite{openai2023chatgpt}. %LLMs have revolutionized natural language processing by demonstrating exceptional reasoning, prediction, and generalization capabilities across various domains. %Their proficiency in few-shot and zero-shot learning scenarios showcases their ability to perform new tasks with minimal task-specific training, highlighting their potential adaptability to diverse applications.
LLMs have revolutionized natural language processing with their exceptional reasoning, prediction, and generalization capabilities, enabling adaptability across diverse applications with minimal task-specific training. 
Despite their success in other fields, the application of LLMs in urban dynamics prediction from spatial-temporal data remains relatively unexplored, with only a few notable attempts. We term these attempts as LLM-based approaches. 
For instance, UrbanGPT~\cite{UrbanGPT} integrates a spatial-temporal dependency encoder with an instruction-tuning paradigm in LLMs. However, it does not consider the inter-correlations among different urban dynamics, limiting its predictive performance in complex, multifaceted urban scenarios. Similarly, methods such as ST-LLM~\cite{ST-LLM}, TPLLM~\cite{TPLLM}, GATGPT~\cite{GATGPT}, and STG-LLM~\cite{STG-LLM} attempt to adapt LLMs to spatial-temporal data. However, these models lack the capability to utilize prompts for prediction guidance and fail to address distributional shifts between training and testing data, reducing their flexibility and adaptability in diverse urban settings.
%These limitations highlight the need for a more comprehensive model that can effectively leverage LLMs' strengths while addressing the unique challenges of spatial-temporal urban prediction.

\noindent\textbf{Our objectives and challenges.} 
Building on the strengths and limitations of NN- and LLM-based approaches, this paper proposes a novel method for adapting LLMs to the spatial-temporal domain for multifaceted urban dynamics prediction, achieving high accuracy while demonstrating strong generalization across diverse urban scenarios, including those with no prior data or unseen conditions.
%By leveraging the reasoning and prediction abilities of LLMs, our objective is to design a model capable of understanding complex spatial-temporal patterns and accurately forecasting multi-faceted urban dynamics,  even in scenarios with no prior data or unseen conditions. 
Achieving this goal requires overcoming below challenges:

% \noindent \underline{\textit{Challenge 1: Data Representation}}: 
% As shown in Figure~\ref{fig:intro}(a), spatial-temporal urban data are inherently complex, with dependencies spanning both spatial and temporal dimensions. Additionally, interactions among multifaceted urban dynamics significantly influence each other. Capturing these complex spatial-temporal dependencies and multifaceted correlations, and transforming such rich information into tokenized representations compatible with LLM is challenging.

\noindent \underline{\textit{Challenge 1: Bridging Spatial-Temporal Data and LLMs}}:
As shown in Figure~\ref{fig:intro}(a), spatial-temporal urban data are inherently complex, exhibiting structured dependencies across space, time, and multiple dynamic factors. However, LLMs are not inherently designed to process such structured, continuous signals, especially time-series data~\cite{tan2024language}. This creates a fundamental representation gap: raw spatial-temporal data must be transformed into discrete, semantically meaningful token representations to be interpretable by LLMs. Bridging this gap is crucial for enabling LLMs to reason over urban dynamics.

\noindent \underline{\textit{Challenge 2: Prompt Design and Fine-Tuning}}: %Designing prompts and fine-tuning strategies to enable LLMs to effectively comprehend spatial-temporal patterns within specific urban contexts presents a significant challenge. 
Spatial and temporal information inherently contain rich semantic details that enhance a model’s understanding of spatial-temporal patterns. Effectively leveraging these details to design prompts for LLM fine-tuning is crucial for enabling LLMs to comprehend spatial-temporal dynamics within specific urban contexts, facilitating reasoning over the data and producing reliable forecasts.%Spatial and temporal information inherently carry valuable semantic details that can enhance the model’s understanding of spatial-temporal patterns. Leveraging these details to craft well-designed prompts to guide 
%LLM finetuning to effectively comprehend spatial-temporal patterns within specific urban contexts is crucial for guiding LLMs to reason over the data and produce reliable forecasts.
%Achieving accurate predictions and strong generalization requires the development of effective prompts and fine-tuning strategies. These strategies must align spatial-temporal embeddings with the LLM’s architecture, enabling the model to reason over complex spatial-temporal interdependencies and deliver reliable forecasts.

\noindent \underline{\textit{Challenge 3: Distributional Shift}}: Urban dynamics often exhibit substantial distributional shifts between training and testing data as shown in Figure~\ref{fig:intro}(b), particularly in zero-shot scenarios where the model is tested in unseen settings with no prior data available. %For instance, testing regions might be completely unseen during training, with distinct road networks and environmental factors resulting in substantially different traffic dynamics. 
These shifts, often reflecting significant differences in geographical characteristics, or temporal patterns compared to the training data, pose a major barrier to generalization and can severely degrade prediction performance.

%Addressing these shifts is crucial to enhance the model’s robustness and ensure accuracy in diverse testing scenarios. Developing mechanisms to adapt the model effectively during testing can mitigate performance degradation and improve zero-shot and few-shot learning performance.

%
%Besides, a prompt and fine-tuning strategy is tailored for MetroMind, guiding the the fine-tuning process and  enabling accurate predictions by leveraging the semantic richness of the spatial-temporal data.
\noindent\textbf{Our UrbanMind.} 
To address these challenges, we propose UrbanMind, a novel multifaceted spatial-temporal LLM that can achieve high prediction accuracy and robust generalization in diverse urban scenarios. 
% UrbanMind incorporates a novel mask-empowered representation learning framework implemented through a novel multifaceted fusion masked autoencoder (Muffin-MAE), which employs specifically designed masking mechanisms to capture complex spatial-temporal dependencies and intercorrelations across multifaceted urban dynamics. 
UrbanMind integrates a Mask-Empowered Representation Learning framework through the novel Muffin-MAE, which employs specialized masking mechanisms to capture complex spatial-temporal dependencies and intercorrelations among multifaceted urban dynamics.
Additionally, it incorporates a tailored prompting and fine-tuning strategy for spatial-temporal data. The prompt encodes spatial-temporal semantic information, guiding the fine-tuning process of the LLM. To generate predictions, a predictor module is attached to the LLM, transforming the high-dimensional latent vectors generated by the LLM into spatial-temporal urban dynamics forecasts.
To address distributional shifts, UrbanMind features a novel test time adaptation strategy with a data reconstructor that operates parallel to the predictor and shares layers with it. During testing, this module adapts to test data by reconstructing the LLM output and fine-tuning the shared layers. 
This ensures improved alignment with the test data distribution, thereby enhancing the generalization capability of the predictor.
%incorporating rich semantic details about the spatial-temporal context of the region and time frame to be predicted. These prompts guide the fine-tuning of the LLM, which is then equipped with a prediction module. This module transforms the high-dimensional latent vectors generated by the LLM into real-world spatial-temporal urban dynamics, enabling accurate predictions while leveraging the semantic richness of the data.
%To address distributional shifts, MetroMind features a novel testing data reconstruction module, operating parallel to the prediction module and sharing layers with it. During testing, this module adapts to new conditions by reconstructing the LLM’s latent vectors using a few epochs of testing data. This quick adaptation process fine-tunes the shared layers, aligning them with the testing distribution. Once adapted, the updated prediction module generates more accurate results, effectively mitigating distributional shifts and enhancing generalization.
\textit{The primary contributions of this paper can be summarized as follows:}
\begin{itemize}[nosep, leftmargin=*]
\item We propose a novel multifaceted spatial-temporal LLM, UrbanMind, which integrates innovative designs to effectively enable a language model to proficiently understand and process the intricate relationships and patterns inherent in multifaceted spatial-temporal data. UrbanMind achieves high prediction accuracy and robust generalization in diverse urban scenarios, including zero-shot cases with no prior data available.

\item UrbanMind incorporates a novel multifaceted fusion masked autoencoder (Muffin-MAE) with advanced masking strategies to generate embeddings that capture both spatial-temporal dependencies and multifaceted intercorrelations, enabling seamless integration with LLMs. Additionally, a semantic-rich prompt and fine-tuning strategy tailored for spatial-temporal data is designed, along with an innovative test time adaptation mechanism that mitigates distributional shifts through a test data reconstructor.

\item We conducted extensive experiments on three different urban dynamics—traffic speed, inflow, and travel demand—in three different cities. The results demonstrate the proposed UrbanMind's exceptional ability to generalize across diverse spatial-temporal learning scenarios and deliver accurate predictions under various conditions, outperforming state-of-the-art baselines, even in unseen regions or scenarios with no prior training data. {\em We have also made our data and code available to the research community ~\footnote{UrbanMind code: \url{https://github.com/Yliu1111/UrbanMind.git}.}}.
\end{itemize}

%% file: Content/overview.tex
\section{Problem Definition}\label{sec:overview}
%In this section, we formally define the urban dynamics prediction problem. %For better clarity and ease of understanding, a summary of the notations used throughout this paper is provided in Table~\ref{tab:notations}.

%Urban dynamics encompass a wide range of elements such as traffic speed, vehicle inflow and outflow, human mobility, \etc. The statistical attributes of these aspects, which exhibit variation across diverse geographical locales and evolve temporally, serve as key indicators of a city's urban dynamic status. 
%This subdivision facilitates a detailed examination of the urban dynamics within specific geographical locations. 
%By considering each grid cell as a distinct entity, our study focuses on understanding how its urban dynamics are shaped by the spatial and temporal characteristics of its neighboring cells.

\noindent\textbf{Definition 1 (Grid Cell $s_{ij}$).}
To achieve a more granular understanding of urban dynamics, a city is partitioned into defined grid cells. For instance, the city can be divided into $J \times J$ grid cells, each with uniform dimensions (\eg, $1 \times 1 , \text{km}^2$). The set of all grid cells is represented as $\mathcal{S} = \{s_{ij}\}$, where $1 \leq i \leq J$ and $1 \leq j \leq J$.

\noindent\textbf{Definition 2 (Target Region $r_{ij}$).}
Each grid cell $s_{ij}$ corresponds to a target region $r_{ij}$. A target region $r_{ij}$ is a square geographic area consisting of $\ell \times \ell$ grid cells, with the cell $s_{ij}$ positioned at its top-left corner. Formally, a target region is represented as $r_{ij} = \langle s_{ij}, \ell \rangle$. The set of all regions is denoted as $\mathcal{R} = \{r_{ij}\}_{1\leq i, j\leq J}$.

%\noindent\textbf{Definition 3 (Urban Dynamics $\bm{X}$).}
%Urban dynamics encompass a wide range of elements, such as traffic speed, vehicle inflow and outflow, human mobility, and more. We represent urban dynamics $\bm{X}$ as a five-dimensional tensor with dimensions $m \times T \times C \times l \times l$, where $m$ is the number of regions in a city, $T$ represents the time steps, $C$ is the number of urban dynamics, and $l \times l$ denotes the size of each region.

%\noindent\textbf{Definition 3 (Urban Dynamics $X$).} Urban dynamics of a region $r$ encompasses a wide range of aspects, such as traffic speed, vehicle inflow and outflow, human mobility, \etc. We represent urban dynamics for a region $r$ with $\bm{X}\in\mathbb{R}^{N \times T \times C \times \ell \times \ell}$ as a five-dimensional tensor, where $N$ is the number of days, $T$ represents the number of time steps per day (\eg, hours), $C$ is the number of urban dynamic aspects that we name as channel number, and $\ell \times \ell$ denotes the size of the region. We further denote a target urban dynamics of interest for a region $r$ as $\bm{X}^k$, \ie, $\bm{X}^k \in\mathbb{R}^{N \times T \times 1 \times \ell \times \ell}$.  

\noindent\textbf{Definition 3 (Urban Dynamics $\mathcal{X}$ and $\mathcal{X}^k$).}  
Urban dynamics of a region \( r \) encompass various aspects, such as traffic speed, vehicle inflow and outflow, and human mobility, among others. We represent the urban dynamics for a region \( r \) as a five-dimensional tensor \(\mathcal{X} \in \mathbb{R}^{N \times T \times C \times \ell \times \ell}\), where \(N\) is the number of days, \(T\) represents the number of time steps per day (e.g., hours), \(C\) denotes the number of urban dynamic aspects, referred to as the number of channels, and \(\ell \times \ell\) specifies the spatial dimensions of the region. We further denote the target urban dynamics of interest for a region \( r \) as \(\mathcal{X}^k\), where \(\mathcal{X}^k \in \mathbb{R}^{N \times T \times 1 \times \ell \times \ell}\), representing a single-channel subset of \(\mathcal{X}\) corresponding to the specific dynamic being analyzed.

%This representation allows us to comprehensively capture the spatial-temporal characteristics of urban dynamics over multiple days and time intervals.

%for a specific grid cell $s$ during a given time interval $t$ to be represented by a scalar value ${x}_t^k$. The collective features of type $k$ from all grid cells within a designated target region $R$ at time $t$ constitute a feature map for that region. This feature map is represented as a matrix $\bm{X}_t^k \in \mathbb{R}^{\ell\times\ell}$, and there are $K$ distinct types of features, thereby we have $1 \leq k \leq K$. 

%\noindent\textbf{Definition 4 (Urban dynamic history $\mathcal{D}^k$).} 
%Urban dynamics observations $\mathcal{D}^k$ include the data for an urban feature $k$ over a span of time, specifically, $\mathcal{D}^k = \{\bm{X}^k_t\}_{t=1}^T$, where $T$ denotes the series length of $\mathcal{D}^k$. For any given time slot $t$, $\mathcal{D}^k_t$ represents the accumulated historical observations up to (and including) that time slot, formalized as $\mathcal{D}^k_t=\{\bm{X}^k_1, \cdots, \bm{X}^k_t \}$.

\noindent\textbf{Problem Definition.} 
%In our paper, we aim to solve the urban dynamics prediction problem, which can be categorized into 2 scenarios:
%
%\noindent\underline{\textit{1. Spatial-Temporal Prediction.}}  
%In this scenario, given the starting urban dynamics data for multiple days, $\mathcal{X}_{\text{start}}$, which represents the urban dynamics for the first $h$ hours of each day across regions $\mathcal{R} = \{R_{ij}\}$, the goal is to predict the subsequent $k$ hours of urban dynamics, $\mathcal{X}_{\text{pred}}$, for these regions over the same days. The spatial-temporal prediction task is formulated as learning a $\theta$-parameterized model:  
%$$
%\mathcal{F}: \mathcal{X}_{\text{pred}} = \mathcal{F}_{\theta}\left(\mathcal{X}_{\text{start}}\right).
%$$  
%
%\noindent\textit{Note:} The model $\mathcal{F}_{\theta}$ should be trained using historical urban dynamics data, $\mathcal{X}_{\text{hist}}$, from the same regions used during testing.
%
Given the historical, multifaceted urban dynamics data $\mathcal{X}^{\text{tr}}$ for training regions denoted as $\mathcal{R}^{\text{tr}} = \{r_{ij}^\text{tr}\}$, the goal is to train a generalizable function $f^*$ that is able to adapt to testing regions $\mathcal{R}^{\text{ts}} = \{r_{ij}^\text{ts}\}$. Specifically, the function $f^*$ is able to predict the urban dynamics for the next $m$ hours in testing regions $\mathcal{R}^{\text{ts}} = \{r_{ij}^\text{ts}\}$, using $h$ hours of prior observations $\mathcal{X}^{\text{ts}}_{\text{start}}$ from these testing regions. 
Formally, the task is expressed as:
$$\hat{\mathcal{X}}^{\text{ts}} = f^*\left(\mathcal{X}^{\text{ts}}_{\text{start}}\right), \text{where } f^*=\arg\min_{f}\mathcal{L}(f, \mathcal{X}^{\text{tr}}),$$
where 
% $\mathcal{X}^{\text{ts}}_{\text{start}}$ represents the urban dynamics data for the first $h$ hours of each day in the testing regions, and 
$\hat{\mathcal{X}}^{\text{ts}}$ denotes the predicted urban dynamics for the following $m$ hours, and $\mathcal{L}$ is the loss function while training the $f$ function.

\noindent\textit{Note:} In the \underline{standard spatial-temporal prediction scenario}, $\mathcal{R}^{\text{ts}} \subseteq \mathcal{R}^{\text{tr}}$. In the 
 \underline{zero-shot prediction scenario}, the testing regions $\mathcal{R}^{\text{ts}}$ must be unseen during the training process and completely distinct from $\mathcal{R}^{\text{tr}}$, \ie, $\mathcal{R}^{\text{ts}}\cap\mathcal{R}^{\text{tr}}=\emptyset$.

%\noindent During training, the model $\mathcal{F}_{\theta}$ is learned by minimizing the Mean Squared Error (MSE) between the predicted and actual urban dynamics for the training regions $\mathcal{R}^{\text{tr}}$:
%$$
%\mathcal{L}(\theta) = \frac{1}{|\mathcal{R}^{\text{tr}}|} \sum_{R_{ij} \in \mathcal{R}^{\text{tr}}} \frac{1}{k} \sum_{t=1}^{k} \left\| \mathcal{F}_{\theta}\left(\mathcal{X}^{\text{tr}}_{\text{start}}\right) - \mathcal{X}^{\text{tr}}_{\text{true}} \right\|^2,
%$$  
%where $\mathcal{X}^{\text{tr}

%% file: Content/method.tex
\section{Methodology}\label{sec:method}
In this section, we introduce UrbanMind, a novel spatial-temporal LLM designed to advance multifaceted urban dynamics prediction performance and enhance adaptability and generalization. It comprises three components, each addressing the challenges outlined in Section~\ref{sec:introduction}:
(1) {\em Mask-Empowered Representation Learning.} 
UrbanMind includes a novel Muffin-MAE with specifically designed masking mechanisms to project urban dynamics into a latent space, effectively capturing complex spatial-temporal dependencies and interdependencies among multifaceted urban dynamics (see Section~\ref{sec:MAE}).
(2) {\em Semantic-Aware Prompting and Fine-Tuning.} UrbanMind designs semantic-aware prompts that encode detailed spatial-temporal contexts. These prompts guide the fine-tuning of the LLM, enabling it to reason over the spatial-temporal embeddings (see Section~\ref{sec:finetuning}). 
(3) {\em Test time adaptation.} UrbanMind incorporates a data reconstructor that quickly adapts to test data by reconstructing the LLM’s output. This enables the model to better align with the test data distribution, enhancing its generalization capability (see Section~\ref{sec:testing}).

\subsection{Mask-Empowered Representation Learning}\label{sec:MAE}
Mask-Empowered Representation Learning is a critical component of UrbanMind. While LLMs are inherently designed to process natural language data, such as text in the form of tokens, they are not equipped to directly handle spatial-temporal data. To leverage LLMs for urban dynamics prediction, it is crucial to transform spatial-temporal urban dynamics into latent representations that LLMs can effectively process and understand, while preserving both the spatial-temporal dependencies and the intercorrelations among multifaceted urban dynamics.

\begin{figure}[t]
    \centering
    %\vspace{-0.4cm}
    \includegraphics[width=0.9\linewidth]{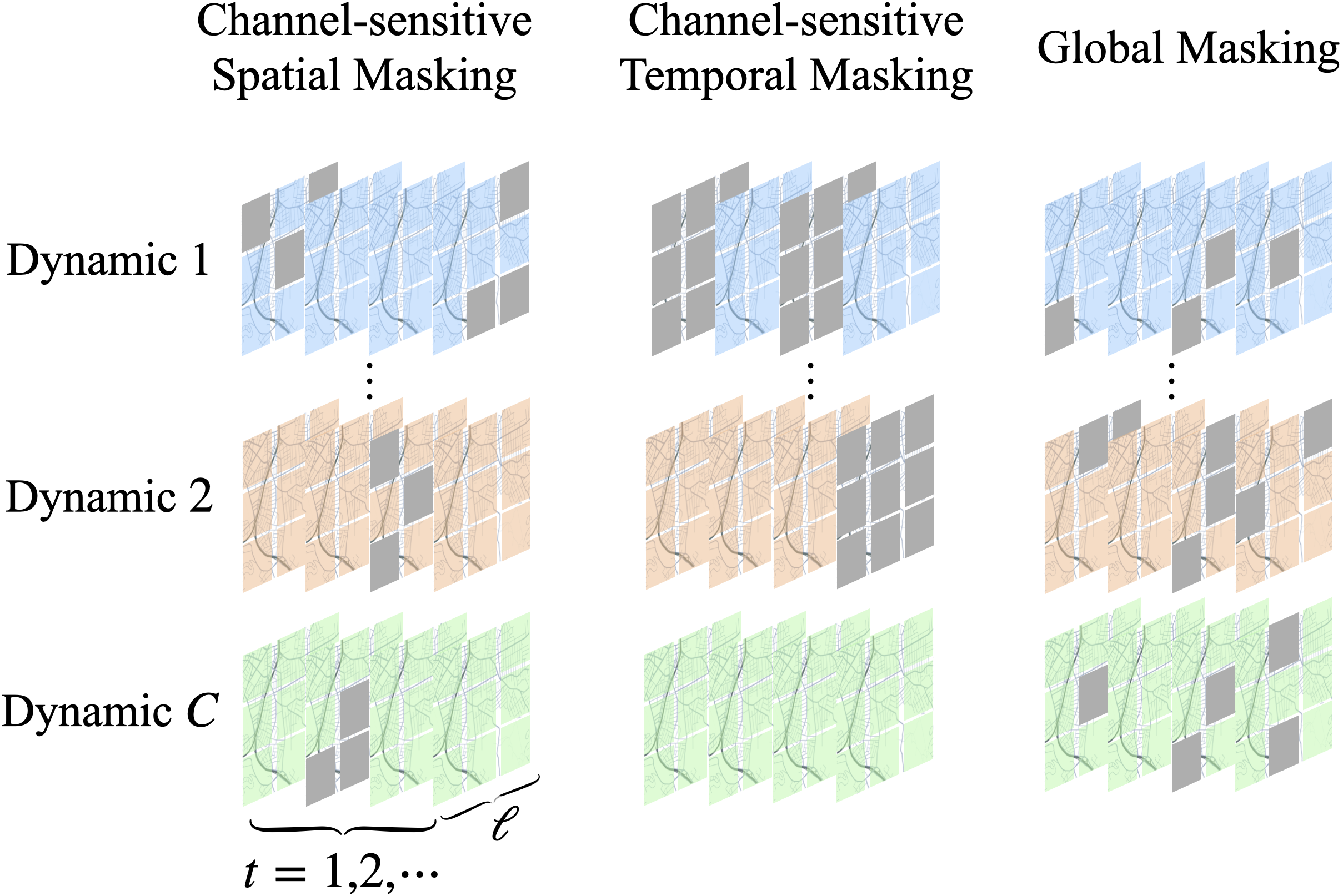}
    \vspace{-0.4cm}
    \caption{Masking strategies. }
    \label{fig:masking}
\vspace{-0.6cm}
\end{figure}

%However, transforming spatial-temporal urban dynamics data into meaningful latent representations is a challenging task. These representations must capture the unique characteristics of spatial-temporal data. For instance, to predict a target urban dynamic, it is crucial to account for its complex spatial-temporal dependencies. Moreover, other related urban dynamics often influence the target urban dynamics, making it necessary to consider the inter-correlations among multi-faceted urban dynamics. As a result, the final latent representation must encapsulate both the spatial-temporal dependencies of the target urban dynamic and the inter-correlations among multi-faceted urban dynamics.

%Muffin-MAE is first trained on multi-faceted urban dynamics to generate embeddings that effectively capture their inter-correlations. The weights of the trained encoder are then partially transferred and fine-tuned on the target urban dynamics to produce target embeddings that preserve the spatial-temporal dependencies of the target dynamic. Finally, the multi-faceted embeddings and target embeddings are combined to form the final spatial-temporal representations, structured as a sequence of tokens suitable for further processing by LLMs.

\begin{figure*}[t]
    \centering
    \includegraphics[width=0.9\linewidth]{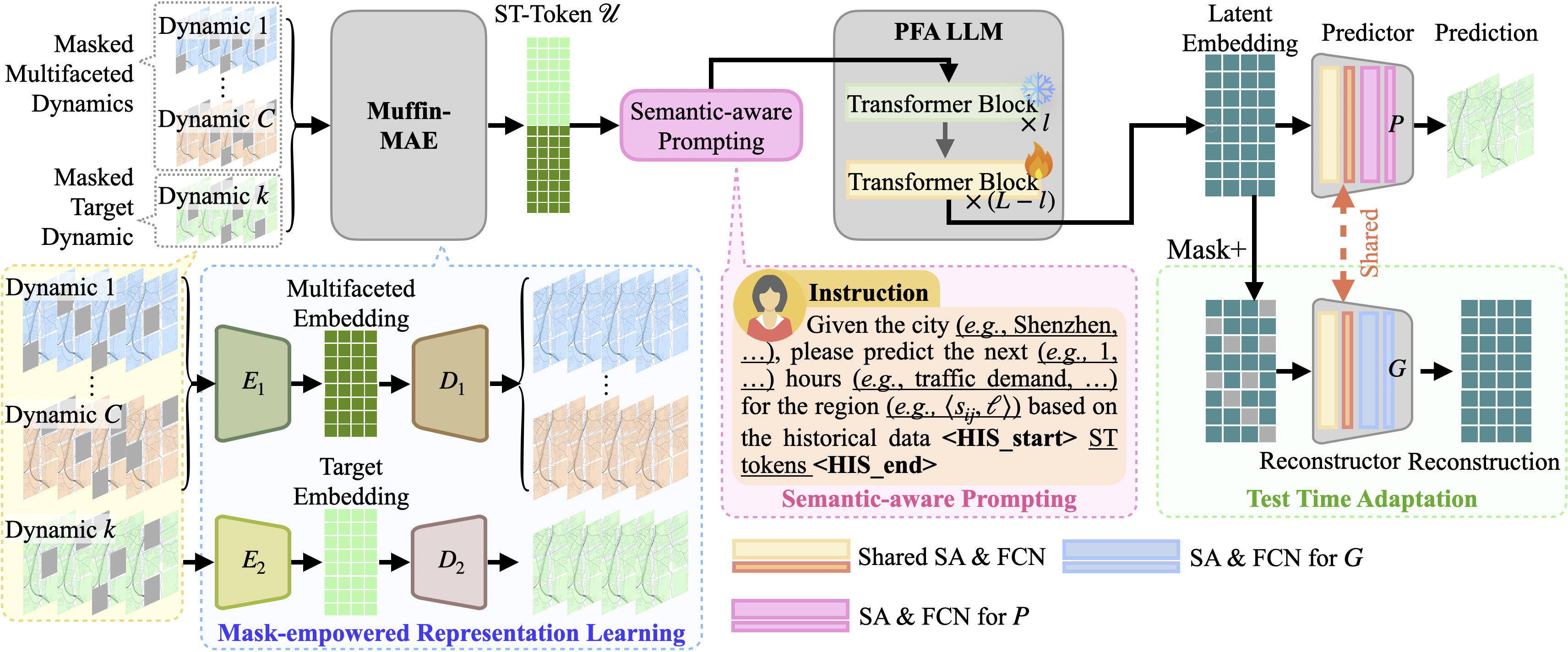}
    \vspace{-0.4cm}
    \caption{UrbanMind Framework. }
    \label{fig:structure}
\vspace{-0.6cm}
\end{figure*}

\noindent\textbf{Multifaceted Fusion MAE (Muffin-MAE).}  
To achieve this, we propose a \underline{mu}lti\underline{f}aceted \underline{f}us\underline{i}o\underline{n} masked autoencoder (Muffin-MAE), a framework with specifically designed masking mechanisms. The Muffin-MAE architecture, illustrated in Figure~\ref{fig:structure}, consists of two masked autoencoders. The first, comprising an encoder \(E_1\) and a decoder \(D_1\), generates multifaceted embeddings by processing multifaceted urban dynamics, effectively capturing their interdependencies. The second, with a separate encoder \(E_2\) and decoder \(D_2\), generates target embeddings for individual urban dynamics, preserving the spatial-temporal dependencies of the target dynamic.

% \begin{figure}[t]
%     \centering
%     \includegraphics[width=1\linewidth]{fig/mae_LLM.png}
%     \caption{Muffin-MAE structure. }
%     \label{fig:mae}
% \end{figure}

Both encoders and decoders consist of multiple convolutional layers, enabling effective processing across spatial and temporal dimensions. The encoder \(E_1\) takes as input a sequence of multifaceted masked urban dynamics data, \(\mathcal{X}^{\text{masked}} = \{\bm{X}_{t}^{\text{masked}}\}_{t=1}^{T}\), where specialized masking mechanisms selectively mask portions of the data. It outputs a sequence of multifaceted latent embeddings, \ie, \(\mathcal{V} = E_1\left(\mathcal{X}^{\text{masked}}\right)\), where \(\mathcal{V} = \{\bm{v}_{t}\}_{t=1}^{T}\) with one embedding generated per time slot. These embeddings are then passed to the decoder \(D_1\), which reconstructs the original multifaceted urban dynamics data, \(\hat{\mathcal{X}} = D_1(\mathcal{V}) = \{\hat{\bm{X}}_{t}\}_{t=1}^{T}\). The objective follows:
\begin{equation}
% &\mathcal{V} = E_1\left(\mathcal{X}^{\text{masked}}\right), \quad \hat{\bm{X}} = D_1\left(\mathcal{V}\right), \label{eq:recons} \\
\mathcal{L}_{\text{Muffin-MAE}} = \frac{1}{T} \sum_{t=1}^{T} \left( \hat{\bm{X}}_{t} - \bm{X}_{t}\right)^2. \label{eq:MAE}
\end{equation}

Once \(E_1\) and \(D_1\) are trained on multifaceted urban dynamics, %the encoder \(E_2\) is initialized with the trained weights from \(E_1\), with shared layers between the two encoders. Separate final layers are attached to \(E_2\) to produce target embeddings for specific urban dynamics. 
as shown in Figure~\ref{fig:structure}, 
\(E_2\) and \(D_2\) operate similarly but process sequences of individual target urban dynamics,  
\(\mathcal{X}^k \in \mathbb{R}^{N \times T \times 1 \times \ell \times \ell}\),  
where the number of channels is set to 1, reflecting the separation of dynamics.

Unlike the masking mechanisms in standard MAE~\cite{ST-MAE}, which primarily rely on space-only or time-only random sampling broadcasted across dimensions, our Muffin-MAE introduces specialized masking strategies designed to capture both spatial and temporal dependencies across different urban dynamics, as illustrated in Figure~\ref{fig:masking} and detailed below.

%\begin{itemize}[nosep, leftmargin=*]
%    \item \textbf{Channel-Sensitive Spatial Masking:}  For a sequence of urban dynamics $\mathcal{X} = \{\bm{X}_{R,t}\}_{t=1}^{T}$, where $\bm{X}_{R,t} \in \mathbb{R}^{C \times l \times l}$, a channel (\ie, one type of urban dynamics) is randomly selected for each $\bm{X}_{R,t}$. Within this channel, $p_s$\% of grid cells are randomly masked without replacement, following a uniform distribution. This mechanism ensures different grid cells are masked across channels, enabling the MAE to effectively capture spatial interdependencies among multi-faceted urban dynamics.

%    \item \textbf{Channel-Sensitive Temporal Masking:}     For the sequence $\mathcal{X} = \{\bm{X}_{R,t}\}_{t=1}^{T}$, $p_t$\% of time steps are randomly sampled from the total $T$. For each sampled time step $\bm{X}_{R,t}^\text{sp}$, a channel is randomly selected, and all grid cells within this channel are masked. This approach allows the MAE to flexibly mask different time steps in different channels, facilitating the capture of temporal interdependencies among multi-faceted urban dynamics.

%    \item \textbf{Global Masking:}      For the sequence $\mathcal{X} = \{\bm{X}_{R,t}\}_{t=1}^{T}$, $p_t$\% of time steps are randomly sampled to form $\{\bm{X}_{R,t}^\text{sp}\}$. For each sampled $\bm{X}_{R,t}^\text{sp}$, $p_s$\% of grid cells across all channels are randomly masked. This global masking mechanism enhances the MAE’s ability to jointly capture spatial and temporal interdependencies across urban dynamics.
%\end{itemize}

\begin{itemize}[nosep, leftmargin=*]
    \item \textbf{Channel-Sensitive Spatial Masking:}  
    For a sequence of urban dynamics \(\mathcal{X} = \{\bm{X}_{r,t}\}_{t=1}^{T}\) at a region $r$ with \(\bm{X}_{r,t} \in \mathbb{R}^{C \times l \times l}\), a channel \(c \in \{1, \dots, C\}\) is randomly selected for each time step \(t\). Within the selected channel \(\bm{X}_{r,t}^c \in \mathbb{R}^{l \times l}\), \(p_s\) of the $\ell\times\ell$ dynamics features are randomly masked with $p_s\in (0,1)$. Let \(\mathcal{M}_{\text{spatial}}\) denote the set of masked feature indices, and \(|\mathcal{M}_{\text{spatial}}| = p_s \cdot l^2\). The masked tensor is given by:
    \[
    \bm{X}_{r,t}^{c,\text{masked}}(i, j) =
    \begin{cases}
    0, & \text{if } (i, j) \in \mathcal{M}_{\text{spatial}}, \\
    \bm{X}_{r,t}^c(i, j), & \text{otherwise}.
    \end{cases}
    \]
    This mechanism ensures that different grid cells of a region are masked across channels, enabling the Muffin-MAE to effectively capture spatial dependencies.

    \item \textbf{Channel-Sensitive Temporal Masking:}  
    For the sequence \(\mathcal{X} = \{\bm{X}_{r,t}\}_{t=1}^{T}\), \(p_t\) of time steps are randomly sampled from \(T\) with $p_t\in(0, 1)$. For each sampled time step \(t^\text{m}\), a channel \(c \in \{1, \dots, C\}\) is randomly selected, and all information corresponding this channel at this region is masked. The masking process for the selected channel is:
    \[
    \bm{X}_{r,t^{\text{m}}}^{c,\text{masked}}(i, j) = 0, \quad \forall (i, j) \in \{1, \dots, l\} \times \{1, \dots, l\}.
    \]
    This approach allows the Muffin-MAE to flexibly mask different time steps in different channels, facilitating the capture of temporal dependencies among multi-faceted urban dynamics.

    % \item \textbf{Global Masking:}  
    % For the sequence \(\bm{X} = \{\bm{X}_{r,t}\}_{t=1}^{T}\), \(p_t\) of time steps are randomly sampled, forming \(\{\bm{X}_{R,t}^\text{sp}\}\). For each sampled time step \(\bm{X}_{R,t}^\text{sp}\), \(p_s\%\) of grid cells across all channels are randomly masked. Let \(\mathcal{M}_{\text{global}}\) denote the set of masked grid cell indices, where \(|\mathcal{M}_{\text{global}}| = p_s \cdot C \cdot l^2\). The masked data is given by:
    \item \textbf{Global Masking:}  
    For the sequence \(\mathcal{X} = \{\bm{X}_{r,t}\}_{t=1}^{T}\), \(p_t\) of time steps are randomly sampled. For each sampled time step \(t^\text{sp}\), \(p_s\) of grid cells across all channels are randomly masked. Let \(\mathcal{M}_{\text{global}}\) denote the set of masked grid cell indices, where \(|\mathcal{M}_{\text{global}}| = p_s \cdot C \cdot l^2\). The masked data is given by:
    \[
    \bm{X}_{r,t}^{\text{masked}}(c, i, j) =
    \begin{cases}
    0, & \text{if } (c, i, j) \in \mathcal{M}_{\text{global}}, \\
    \bm{X}_{r,t}(c, i, j), & \text{otherwise}.
    \end{cases}
    \]
    Global masking mechanism enhances Muffin-MAE to capture both spatial and temporal dependencies across urban dynamics.
\end{itemize}

\noindent\textbf{Multifaceted and Target Embeddings.}  
To obtain the multifaceted embeddings from the multifaceted urban dynamics \(\mathcal{X}\), we train the encoder \(E_1\) and decoder \(D_1\) in Muffin-MAE using all three masking mechanisms. Upon completion of training, the encoder generates embeddings \(\mathcal{V}\), effectively preserving the inter-correlations among the related multi-faceted urban dynamics.
To derive target embeddings \(\mathcal{V}^k = \{\bm{v}_{r,t}^k\}_{t=1}^{T}\) for a sequence of target urban dynamics \(\mathcal{X}^k\), we train the encoder \(E_2\) and the decoder \(D_2\) using the target urban dynamics. %This layer-sharing design leverages \(E_1\)'s ability to capture rich spatial-temporal patterns across various dynamics, making the training process of \(E_2\) both efficient and effective for capturing the spatial-temporal dependencies of the target urban dynamics. 
During this training process, all three masking mechanisms are again applied to the target urban dynamics, with the channel number set to 1.

\noindent \textbf{Spatial-Temporal Token Generation.}
To form the final tokens for a sequence of urban dynamics, we combine the multifaceted embeddings \(\mathcal{V} = \{\bm{v}_{r,t}\}_{t=1}^{T}\) with the target embeddings \(\mathcal{V}^k = \{\bm{v}_{r,t}^k\}_{t=1}^{T}\). This combination is achieved by concatenating the embeddings along the feature dimension for each time step as \(\mathcal{U} = \{\bm{u}_{r,t}\}_{t=1}^{T}\), where
$\bm{u}_{r,t} = \text{concat}(\bm{v}_{r,t}, \bm{v}_{r,t}^k)$. Here, \(\bm{u}_{r,t}\) represents the final token for time step \(t\), encapsulating both the inter-correlations among the related multifacet dynamics and the spatial-temporal dependencies of the target urban dynamics. These tokens, structured as \(\bm{u}_{r,t} \in \mathbb{R}^{d_v + d_k}\), are well-suited for processing by LLMs.

\subsection{Semantic-Aware Prompting and Fine-Tuning}\label{sec:finetuning}

\textbf{Semantic-Aware Prompting Design.}
In spatial-temporal prediction tasks, both temporal and spatial information carry essential semantic details that contribute to the model’s understanding of complex patterns within specific contexts. %Temporal variations, such as traffic speed fluctuations during rush hours versus non-rush hours, and spatial differences, such as distinct traffic patterns in commercial versus residential areas, highlight the importance of integrating both dimensions. 
To leverage these insights, we represent temporal and spatial information as prompt instruction text, enabling large language models to process this data effectively through their advanced text comprehension capabilities.

In the UrbanMind framework, we incorporate time, regional, and task-specific information into the instruction input for the large language model. Temporal information includes data from the previous hours of a day and the target hours whose urban dynamics need to be predicted. Regional information specifies the city, the coordinates of the top-left grid cell $s_{ij}$ within the target region, and the region’s side length $\ell$. Additionally, the task name defines the specific urban dynamics to be predicted. By encapsulating these elements in a structured prompt, UrbanMind effectively identifies and assimilates spatial-temporal patterns across diverse regions, time frames, and urban dynamics as is illustrated in Figure~\ref{fig:structure}. %This comprehensive prompt design enhances the model’s ability to reason within complex spatial-temporal contexts, enabling its generalization capabilities. 
% An illustration of the instructional prompt design is provided in Figure~\ref{fig:mae}.

\noindent\textbf{LLM Fine-Tuning Strategy.}  
To adapt the LLM for spatial-temporal urban dynamics prediction, we employ the Partially Frozen Attention (PFA) mechanism~\cite{ST-LLM}. Consider an LLM (we use LLaMA3.2~\cite{dubey2024llama} in this paper) with \(L\) transformer layers, where each transformer layer \(\text{TFM}_i\) (\(i = 1, \dots, L\)) undergoes the following transformation:
\[
\bm{e}^{(i)} = \text{LayerNorm}(\bm{e}^{(i-1)} + \text{SA}(\bm{e}^{(i-1)})),
\]
where \(\bm{e}^{(i)}\) represents the hidden state produced by the \(i\)-th layer of the LLM, and \(\text{SA}(\cdot)\) denotes the self-attention operation. 

To efficiently fine-tune the model, we divide the transformer layers into two groups: frozen layers \(\text{TFM}^\text{fr} = \{\text{TFM}^{(1)}, \dots, \text{TFM}^{(l)}\}\) and trainable layers \(\text{TFM}^\text{tr} = \{\text{TFM}^{(l+1)}, \dots, \text{TFM}^{(L)}\}\). The parameters of the frozen layers \(\text{TFM}^\text{fr}\) remain fixed during fine-tuning, preserving the pretrained knowledge of the LLM. The trainable layers \(\text{TFM}^\text{tr}\) process the output from the frozen layers, where only the query matrices \(\mathbf{W}_q\) in the self-attention mechanism are updated to learn task-specific spatial-temporal dependencies. In contrast, the key and value matrices \(\mathbf{W}_k\) and \(\mathbf{W}_v\) remain frozen to retain the pretrained relationships encoded in the model.
This two-stage fine-tuning process leverages the generalization capabilities of the frozen layers while allowing the trainable layers to effectively adapt to the spatial-temporal characteristics of urban dynamics.
%This two-stage fine-tuning process leverages the pretrained capabilities of the frozen layers while adapting the trainable layers to effectively model spatial-temporal urban dynamics.

%
%During fine-tuning, the input prompt is processed as follows: \textit{(1) Frozen Layers} $\text{TFM}^\text{fr}$ process the input sequence while keeping the parameters unchanged. 
%Each transformer layer $\text{TFM}_i$ with $i=1,\cdots,l$ goes through the following transformation:
%\[\bm{e}^{(i)} = \text{LayerNorm}(\bm{e}^{(i-1)} + \text{SA}(\bm{e}^{(i-1)})),
%\]
%where \(\bm{e}^{(i)}\) represents the hidden state produced by the $i$-th layer of LLM, and \(\text{SA}(\cdot)\) is the self-attention operation. 
%
%\textit{(2) Trainable Layers} $\text{TFM}^\text{tr}$ also consist of a self-attention operation following a layer normalization in each layer. They process the output from the frozen layers and their query matrix \(\mathbf{W}_q\) in the self-attentions are updated to learn task-specific spatial-temporal dependencies. 
%In contrast, the key and value matrices \(\mathbf{W}_k\) and \(\mathbf{W}_v\) remain frozen during fine-tuning to preserve the relationships encoded in the model. 
% For each layer \(l \in \mathcal{L}_t\), the transformation is given by:
% \[
% \bm{e}_{R,t}^{(l)} = \text{LayerNorm}(\bm{e}_{R,t}^{(l-1)} + \text{SA}(\bm{e}_{R,t}^{(l-1)})),
% \]
% where \(\mathbf{W}_q\), \(\mathbf{W}_k\), and \(\mathbf{W}_v\) are parameters of the self-attention operation \(\text{SA}(\cdot)\). 

\noindent\textbf{Spatial-Temporal Predictor Module.}
The LLM's output—a high-dimensional sequence of latent embeddings—must be transformed into numerical values representing predicted urban dynamics. To realize this goal, we attach a spatial-temporal predictor module \(P\) to the LLM. This module consists of self-attention layers and fully connected layers (in Figure~\ref{fig:structure}), enabling the transformation of LLM-generated embeddings into structured urban dynamics predictions.

The spatial-temporal prediction module \(P\) processes the sequence of embeddings \(\mathcal{E} = \{\bm{e}\}\) generated by the LLM based on the prompt using $h$ hours of prior observations and predicts the urban dynamics \(\hat{\mathcal{X}}^k = \{\hat{\bm{X}}_{r,t}^k\}_{t=h+1}^{h+m}\) for the subsequent $m$ hours, where $\hat{\mathcal{X}}^k = P(\mathcal{E})$.
To optimize the prediction, we minimize the Mean Squared Error (MSE) loss for daily sequential urban dynamics data: 
\begin{equation}
\mathcal{L}_{\text{pred}} = \frac{1}{m} \sum_{t=h+1}^{h+m} \left\| \tilde{\bm{X}}_{r,t}^k - \bm{X}_{r,t}^k \right\|^2,
\label{eq:LLM}
\end{equation}
where \(\hat{\bm{X}}_{r,t}^k\) represents the predicted urban dynamics, and \(\bm{X}_{r,t}^k\) denotes the corresponding ground truth values.

\subsection{Test Time Adaptation}\label{sec:testing}
Although LLMs exhibit strong generalizability, they are primarily designed for natural language tasks and generalization across text-related domains. In the spatial-temporal domain, testing data often presents distinct patterns from training data, especially in zero-shot scenarios where unseen regions and varying traffic dynamics introduce significant distributional shifts.
To address this, we propose a test time adaptation strategy to enhance the generalization and adaptability of LLMs during testing. At its core is a novel test data reconstructor 
\( G \), which works in parallel with the predictor while sharing several self-attention layers as is shown in Figure~\ref{fig:structure}.

During testing, the LLM processes the prompt describing the test region (potentially unseen) and generates a latent embedding sequence \(\mathcal{E} = \{\bm{e}\}\) as input to the predictor. 
Before directly passing \(\mathcal{E}\) to the predictor, we randomly mask \(p\) of the elements (with $p\in(0, 1)$ as the masking ratio) in $\mathcal{E}$ to introduce stochasticity and encourage robustness. The reconstructor  \( G \) then recovers the masked elements through a quick adaptation process, performing a few epochs of updates.
The masking process %for the sequence of latent embeddings  \(\mathcal{E} = \{\bm{e}_{r,t}\}_{t=1}^{h}\), where each \(\bm{e}_{r,t} \in \mathbb{R}^d\), 
works as follows: A binary mask vector \(\bm{m}_i \in \{0, 1\}^d\) is generated for each \(\bm{e}_i\in\mathbb{R}^d\). The indices to be masked, \(\mathcal{M} \subseteq \{1, \dots, d\}\), are randomly sampled with a uniform distribution such that \(|\mathcal{M}| = p \cdot d\).], \ie, %The elements of \(\bm{m}_i\) are given by:
%\[
%m_{i,i} =
%\begin{cases} 
%0, & \text{if } j \in \mathcal{M}, \\
%1, & \text{otherwise}.
%\end{cases}
%\]
% 
% The masked embedding \(\bm{e}_i^{\text{masked}}\) is computed as:
\begin{equation}
\bm{e}_i^{\text{masked}} = \bm{e}_i \odot \bm{m}_i,
\label{eq:testing_mask}
\end{equation}
where \(\odot\) denotes element-wise multiplication. The sequence of masked embeddings is then represented as \(\mathcal{E}^{\text{masked}} = \{\bm{e}_i^{\text{masked}}\}\).
The reconstructor \( G \) reconstructs the entire sequence of latent embeddings \(\mathcal{E} = \{\bm{e}\}\) from the masked sequence \(\mathcal{E}^{\text{masked}}\) with loss: 
\begin{equation}
\mathcal{L}_{\text{recon}} = \frac{1}{n} \sum_{i=1}^{n} \left\| G(\bm{e}_i^{\text{masked}}) - \bm{e}_i \right\|^2,
\label{eq:testing}
\end{equation}
where \(n\) is the number of embeddings, \(\bm{e}_i\) is the original embedding, and \(G(\bm{e}_i^{\text{masked}})\) is the reconstructed embedding produced by the reconstructor \( G \).
This reconstruction process allows \( G \) to quickly adapt to new data and regional conditions, fine-tuning the shared layers to better align with the test data. Once adaptation is complete, the updated shared layers enable the predictor to generate more accurate results for testing scenarios. This approach effectively mitigates distributional shifts, enhancing both generalization and adaptability in diverse spatial-temporal contexts\cite{sun2023learning,sun2024learning}.
The detailed algorithm for UrbanMind can be found in Algorithm~\ref{alg:A}.

\renewcommand{\algorithmicrequire}{ \textbf{Input:}}
\renewcommand{\algorithmicensure}{ \textbf{Output:}} 
%\vspace{-0.28cm}
\begin{algorithm}[h]
\caption{Algorithm for UrbanMind}
\label{alg:A}
\begin{algorithmic}[1]
\REQUIRE Target urban dynamics $\mathcal{X}^k$, related multifaceted dynamics $\mathcal{X}$, initialized Muffin-MAE encoders $E_{\phi_1}$ and $E_{\phi_2}$, decoders $D_{\psi_1}$ and $D_{\psi_2}$, predictor $P_\theta$, reconstructor $G_\tau$, and pretrained LLM $Q_\omega$.
\ENSURE Well-trained $E_{\phi_1}$, $E_{\phi_2}$, $D_{\psi_1}$, $D_{\psi_2}$, $Q_\omega$, $P_\theta$ and $G_\tau$.
\STATE {\textbf{Step 1: Mask-Empowered Representation Learning}:}
\FOR{iteration $i =1,2,3,\cdots$}
    \STATE {Sample a batch of $\mathcal{X}$ and $\mathcal{X}^k$.}
    \STATE {Train $E_{\phi_1}$, $E_{\phi_2}$, $D_{\psi_1}$ and $D_{\psi_2}$ with Eq.~\eqref{eq:MAE}.}
\ENDFOR
    \STATE {Get the multifaceted embeddings $\mathcal{V}$ with $E_{\phi_1}$ and the target embeddings $\mathcal{V}^k$ with $E_{\phi_2}$.}
%\FOR{iteration $i =1,2,3,\cdots$}
%    \STATE {Sample a batch of $\mathcal{X}^k$.}
%    \STATE {Train encoder $E_\phi$ and decoder $D_\psi$ with Eq.~\eqref{eq:MAE}.}
%\ENDFOR
    %\STATE {Get the target embeddings $\mathcal{V}^k$ using $E_\phi$.}
    \STATE {Get the final tokens by $\mathcal{U} = \text{concat}(\mathcal{V}, \mathcal{V}^k)$.}
\STATE {\textbf{Step 2: Semantic-Aware Prompting and
Fine-Tuning}:}
\FOR{iteration $=1,2,3,\cdots$}
    \STATE {Combine $\mathcal{U}$ with text descriptions as prompts.}   
    \STATE {Fine-tune $Q_\omega$ and update $P_\theta$ with prompts using Eq.\eqref{eq:LLM}.}
\ENDFOR
\STATE {\textbf{Step 3: Test Time Adaptation and Final Prediction}:}
\FOR{iteration $i = 1, 2, 3, \cdots$}
    \STATE Sample a testing sequence $\mathcal{X}^{\text{ts}}$, and produce the a latent embedding sequence \(\mathcal{E}\) using $Q_\omega$.
    \STATE Get \(\mathcal{E}^{\text{masked}}\) with Eq.\eqref{eq:testing_mask} as the input of $G_\tau$.
    \STATE Update $G_\tau$ with Eq.\eqref{eq:testing}. 
\ENDFOR
\STATE Produce final prediction with $P_\theta(\mathcal{X}^{\text{ts}})$.
\end{algorithmic}
\end{algorithm}

%% file: Content/experiment.tex
\begin{table*}[ht]
\centering
\tiny
\setlength{\tabcolsep}{1pt} % 调整列间距
\renewcommand{\arraystretch}{1.0} % 调整行间距
\resizebox{\textwidth}{!}{ % 调整表格宽度
\begin{tabular}{ccccccccccccccc}
\toprule
\multirow{2}{*}{\textbf{City}} & \multirow{2}{*}{\textbf{Dynamics}} & \multirow{2}{*}{\textbf{Metrics}} & \multicolumn{12}{c}{\textbf{Models}} \\
\cmidrule(lr){4-15}
 &  &  & DYffusion & TGC-LSTM & GCRN & GAGCN & GATGPT & GCNGPT & ST-LLM & TPLLM & LLaMA3 & STG-LLM & UrbanGPT & UrbanMind \\
\midrule

\multirow{6}{*}{Shenzhen} 
& \multirow{2}{*}{Speed}  & MAE  & \textbf{0.131} & 0.196 & 0.196 & 0.361 & 0.139 & 0.159 & 0.194 & 0.166 & 0.256 & 0.194 & 0.132 & \textbf{0.131} \\
         &         & RMSE & 0.211 & 0.263 & 0.265 & 0.485 & 0.205 & 0.225 & 0.290 & 0.229 & 0.443 & 0.274 & 0.201 & \textbf{0.194} \\
         
& \multirow{2}{*}{Inflow} & MAE  & 0.128 & 0.235 & 0.324 & 0.371 & 0.128 & 0.137 & 0.153 & 0.165 & 0.269 & 0.217 & 0.137 & \textbf{0.127} \\
         &         & RMSE & 0.251 & 0.364 & 0.489 & 0.542 & 0.250 & 0.254 & 0.265 & 0.255 & 0.460 & 0.382 & \textbf{0.249} & \textbf{0.249} \\
         
& \multirow{2}{*}{Demand} & MAE  & 0.149 & 0.228 & 0.324 & 0.464 & 0.141 & 0.153 & 0.207 & 0.178 & 0.223 & 0.202 & 0.144 & \textbf{0.138} \\
         &         & RMSE & 0.339 & 0.378 & 0.504 & 0.644 & 0.270 & 0.275 & 0.301 & 0.278 & 0.381 & 0.377 & 0.271 & \textbf{0.269} \\

\midrule

\multirow{6}{*}{Xi'an} 
& \multirow{2}{*}{Speed} & MAE  & 0.216 & 0.201 & 0.230 & 0.481 & 0.268 & 0.138 & 0.220 & 0.198 & 0.294 & 0.208 & 0.139 & \textbf{0.137} \\
         &         & RMSE & 0.253 & 0.227 & 0.263 & 0.314 & 0.296 & 0.183 & 0.323 & 0.265 & 0.417 & 0.228 & 0.192 & \textbf{0.162} \\
         
 & \multirow{2}{*}{Inflow} & MAE  & 0.386 & 0.263 & 0.419 & 0.360 & 0.300 & 0.142 & 0.231 & 0.330 & 0.356 & 0.334 & 0.170 & \textbf{0.114} \\
         &         & RMSE & 0.552 & 0.392 & 0.452 & 0.422 & 0.341 & 0.194 & 0.340 & 0.460 & 0.530 & 0.414 & 0.237 & \textbf{0.173} \\
         
 & \multirow{2}{*}{Demand} & MAE  & 0.330 & 0.291 & 0.291 & 0.303 & 0.337 & 0.210 & 0.195 & 0.333 & 0.279 & 0.263 & 0.185 & \textbf{0.160} \\
         &         & RMSE & 0.314 & 0.430 & 0.358 & 0.416 & 0.384 & 0.341 & 0.290 & 0.441 & 0.435 & 0.285 & 0.284 & \textbf{0.273} \\

\midrule

\multirow{6}{*}{Chengdu} 
& \multirow{2}{*}{Speed} & MAE  & 0.201 & 0.157 & 0.202 & 0.460 & 0.253 & 0.135 & 0.302 & 0.185 & 0.289 & 0.175 & 0.133 & \textbf{0.120} \\
         &         & RMSE & 0.240 & 0.204 & 0.251 & 0.501 & 0.282 & 0.190 & 0.349 & 0.238 & 0.463 & 0.194 & 0.191 & \textbf{0.166} \\
         
 & \multirow{2}{*}{Inflow} & MAE  & 0.309 & 0.328 & 0.550 & 0.247 & 0.334 & 0.191 & 0.217 & 0.348 & 0.349 & 0.308 & 0.203 & \textbf{0.187} \\
         &         & RMSE & 0.520 & 0.454 & 0.646 & 0.361 & 0.379 & 0.243 & 0.324 & 0.469 & 0.480 & 0.389 & 0.262 & \textbf{0.240} \\
         
& \multirow{2}{*}{Demand} & MAE  & 0.376 & 0.284 & 0.452 & 0.297 & 0.259 & 0.171 & 0.169 & 0.290 & 0.264 & 0.356 & 0.160 & \textbf{0.125} \\
         &         & RMSE & 0.406 & 0.405 & 0.486 & 0.416 & 0.301 & 0.230 & 0.233 & 0.388 & 0.481 & 0.381 & 0.216 & \textbf{0.202} \\

\bottomrule
\end{tabular}
}
\caption{Zero-shot Prediction: Average performance for speed, inflow, and demand predictions across 3 cities. In this zero-shot case, all testing regions are unseen during training.}
\label{tab:zero_shot_baselines}
\vspace{-0.7cm}
\end{table*}

\begin{figure*}[ht]
    \centering
    % \begin{minipage}{.47\columnwidth}
        \includegraphics[width=0.9\linewidth]{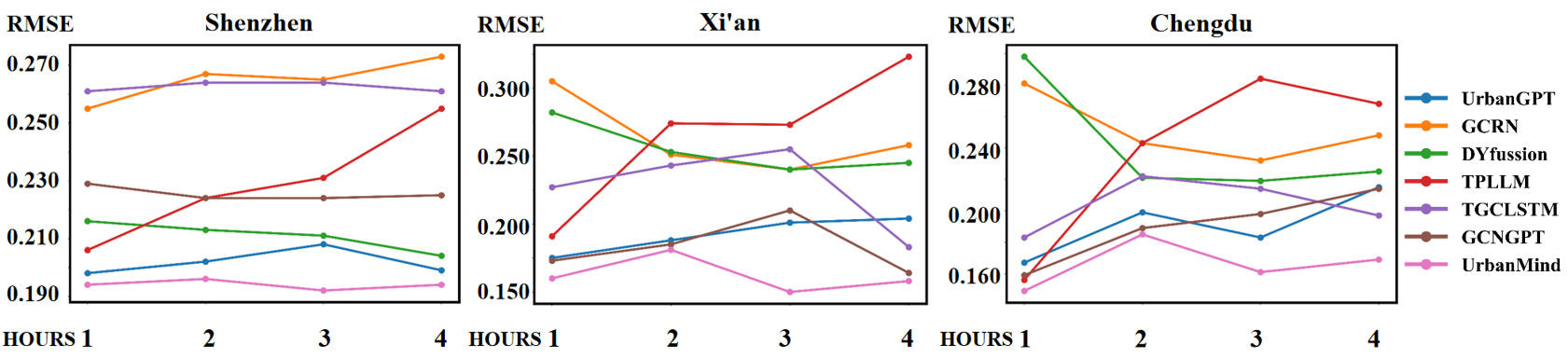} 
    % \end{minipage}\hfill
    \vspace{-10pt}
    \caption{Zero-shot Prediction: RMSE for traffic speed prediction over 4 hours in Shenzhen, Xi'an, and Chengdu.}
    \label{fig:main}
\vspace{-5mm}
\end{figure*}

\section{Experiment}\label{sec:experiment}
In experiments, we aim to measure the effectiveness of our UrbanMind across different datasets and in different urban scenarios. We will answer the following questions with extensive experiments: (1) Can our UrbanMind outperform in zero-shot spatial-temporal prediction tasks for different urban dynamics? (2)  Can our UrbanMind outperform in standard spatial-temporal prediction tasks for different urban dynamics? (3) Is each component in UrbanMind effective when performing spatial-temporal predictions? (4) How do different hyperparameters affect performance?

\subsection{Dataset and Experiment Descriptions}
\textbf{Data description.} 
In our experiments, we use three distinct urban dynamics datasets—travel demand, traffic speed, and inflow—from multiple cities, including Chengdu, Xi'an, and Shenzhen. 
For each city, the entire area is partitioned into equal-sized grid cells with a region size of \(10 \times 10\). %Traffic speed and inflow are used to measure the traffic status in each grid cell. Travel demand captures the level of movement within a grid cell, quantified by the number of trips, such as taxi pickups and drop-offs, occurring during a given time slot. 
In Shenzhen, each dataset is processed to yield dimensions of \((162, 12, 63, 10, 10)\), where 162 represents the number of days, 12 corresponds to the twelve one-hour time slots per day, and 63 indicates the number of regions. For Chengdu and Xi'an, each type of urban dynamics is structured with dimensions \((30, 12, 4, 10, 10)\), where 30 denotes the number of days, 12 indicates the time slots per day and 4 represents the number of regions. All the data and regions in all datasets are divided into training and testing. More details can be found in the Appendix.

\noindent\textbf{Task Description.}  
To validate the prediction accuracy and generalizability of UrbanMind, we conduct experiments on two tasks:  \underline{\textit{(1) Zero-shot Prediction.}} We evaluate the model's generalizability by predicting future spatial-temporal data from unseen regions in three cities for all three urban dynamics types.
\underline{\textit{(2) Standard Prediction.}} We assess the model's prediction accuracy by predicting future urban dynamics for the same regions included in the training set.

\begin{table*}[ht]
\centering
\tiny
\setlength{\tabcolsep}{1pt} % 调整列间距
\renewcommand{\arraystretch}{1.0} % 调整行间距
\resizebox{\textwidth}{!}{ % 调整表格宽度
\begin{tabular}{ccccccccccccccc}
\toprule
\multirow{2}{*}{\textbf{City}} & \multirow{2}{*}{\textbf{Dynamics}} & \multirow{2}{*}{\textbf{Metrics}} & \multicolumn{12}{c}{\textbf{Models}} \\
\cmidrule(lr){4-15}
 &  &  & DYffusion & TGC-LSTM & GCRN & GAGCN & GATGPT & GCNGPT & ST-LLM & TPLLM & LLaMA3 & STG-LLM & UrbanGPT & UrbanMind \\
\midrule

\multirow{6}{*}{Shenzhen} 
& \multirow{2}{*}{Speed}  & MAE  & 0.175 & 0.187 & 0.202 & 0.307 & 0.155 & 0.161 & 0.201 & 0.225 & 0.261 & 0.167 & 0.147 & \textbf{0.130} \\
         &         & RMSE & 0.239 & 0.218 & 0.235 & 0.515 & 0.203 & 0.230 & 0.281 & 0.248 & 0.454 & 0.245 & 0.205 & \textbf{0.200} \\

& \multirow{2}{*}{Inflow} & MAE  &0.272 &  0.142 & 0.132 & 0.182 & 0.136 & 0.301 & 0.286 & 0.181 & 0.268 & 0.240 & 0.128 & \textbf{0.123} \\
         &         & RMSE & 0.443 & 0.207 & 0.189 & 0.271 & 0.235 & 0.389 & 0.381 & 0.211 & 0.450 & 0.400 & 0.190 & \textbf{0.185} \\
         
& \multirow{2}{*}{Demand} & MAE  & 0.285 & 0.166 & 0.168 & 0.338 & 0.154 & 0.228 & 0.223 & 0.178 & 0.236 & 0.371 & 0.137 & \textbf{0.133} \\
         &         & RMSE & 0.389 & 0.257 & 0.220 & 0.518 & 0.275 & 0.343 & 0.325 & 0.222 & 0.380 & 0.464 & 0.210 & \textbf{0.199} \\

\midrule

\multirow{6}{*}{Xi'an} 
& \multirow{2}{*}{Speed}& MAE  & 0.210 & 0.156 & 0.225 & 0.421 & 0.304 & 0.291 & 0.425 & 0.270 & 0.317 & 0.183 & 0.134 & \textbf{0.114} \\
         &         & RMSE & 0.266 & 0.201 & 0.264 & 0.524 & 0.358 & 0.415 & 0.520 & 0.348 & 0.516 & 0.220 & 0.190 & \textbf{0.186} \\

 & \multirow{2}{*}{Inflow} & MAE  & 0.304 & 0.254 & 0.430 & 0.341 & 0.311 & 0.298 & 0.289 & 0.334 & 0.363 & 0.350 & 0.214 & \textbf{0.124} \\
         &         & RMSE &  0.533 &0.384 & 0.512 & 0.546 & 0.367 & 0.437 & 0.368 & 0.442 & 0.551 & 0.383 & 0.330 & \textbf{0.181} \\

 & \multirow{2}{*}{Demand} & MAE  & 0.259 & 0.329 & 0.378 & 0.377 & 0.381 & 0.401 & 0.199 & 0.382 & 0.327 & 0.359 & 0.253 & \textbf{0.159} \\
         &         & RMSE &  0.439& 0.466 & 0.495 & 0.501 & 0.431 & 0.525 & 0.281 & 0.489 & 0.513 & 0.401 & 0.407 & \textbf{0.273} \\

\midrule

\multirow{6}{*}{Chengdu} 
& \multirow{2}{*}{Speed} & MAE  &  0.193 &0.137 & 0.216 & 0.516 & 0.289 & 0.292 & 0.448 & 0.229 & 0.293 & 0.152 & 0.147 & \textbf{0.119} \\
         &         & RMSE & 0.244 & 0.182 & 0.255 & 0.627 & 0.331 & 0.396 & 0.528 & 0.289 & 0.480 & 0.184 & 0.170 & \textbf{0.165} \\
         
 & \multirow{2}{*}{Inflow} & MAE  &0.361  & 0.287 & 0.315 & 0.320 & 0.234 & 0.251 & 0.326 & 0.285 & 0.343 & 0.397 & 0.267 & \textbf{0.153} \\
         &         & RMSE &0.521 &  0.381 & 0.451 & 0.433 & 0.287 & 0.338 & 0.409 & 0.387 & 0.490 & 0.478 & 0.389 & \textbf{0.191} \\

& \multirow{2}{*}{Demand} & MAE  & 0.306 & 0.271 & 0.304 & 0.378 & 0.251 & 0.194 & 0.236 & 0.257 & 0.289 & 0.316 & 0.229 & \textbf{0.153} \\
         &         & RMSE & 0.491 & 0.390 & 0.402 & 0.493 & 0.392 & 0.236 & 0.321 & 0.358 & 0.479 & 0.375 & 0.340 & \textbf{0.187} \\

\bottomrule
\end{tabular}
}
\caption{Standard Prediction: Average performance of speed, inflow, and demand prediction across 3 cities. Historical data for testing regions were included during training, and future dynamics are predicted.}
\label{tab:same_region_baselines}
\vspace{-0.7cm}
\end{table*}

\begin{figure*}[t]
    \centering
    % \begin{minipage}{.47\columnwidth}
        \includegraphics[width=0.9\linewidth]{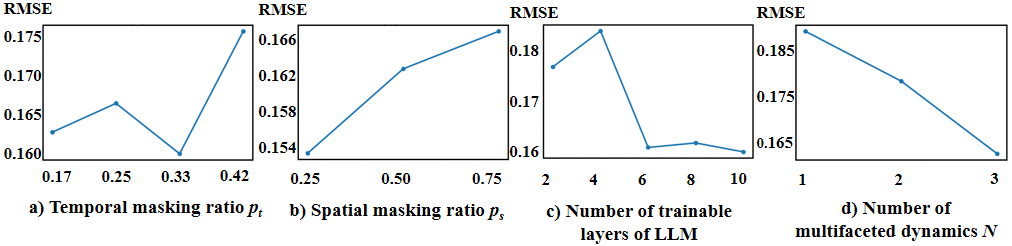} 
    % \end{minipage}\hfill
    \vspace{-4mm}
    \caption{Impact of Hyperparameters on RMSE Performance for Traffic Speed Prediction in Xi'an.}
    \label{fig:hyper}
\vspace{-5mm}
\end{figure*}

\subsection{Baselines}
We compare UrbanMind against state-of-the-art urban dynamics prediction models, including LLM-based approaches:  
\textbf{GATGPT}~\cite{chen2023gatgpt}, \textbf{GCNGPT}~\cite{chen2023gatgpt,liu2024spatial}, \textbf{ST-LLM}~\cite{liu2024spatial}, \textbf{TPLLM}~\cite{ren2024tpllm}, \textbf{UrbanGPT}~\cite{li2024urbangpt}, and \textbf{STG-LLM}~\cite{liu2024can}. These models integrate LLMs with CNNs or GCNs to enhance urban dynamics prediction. 
% Additionally, since we use \textbf{LLaMA3}~\cite{dubey2024llama} as the foundation model for UrbanMind, we evaluate LLaMA3 as a baseline to demonstrate the infeasibility of directly applying LLMs to urban dynamics prediction. 
Additionally, we use \textbf{LLaMA3.2}~\cite{dubey2024llama} as UrbanMind’s foundation model, referred to as \textbf{LLaMA3}. It is also evaluated as a baseline to show the limits of directly applying LLMs to urban dynamics prediction.
Beyond these, we also benchmark against most recent and effective neural network-based models, including \textbf{DYffusion}~\cite{cachay2024dyffusion}, \textbf{TGC-LSTM}~\cite{cui2019traffic}, \textbf{GCRN}~\cite{seo2018structured}, and \textbf{GAGCN}~\cite{xia2022short}. More details on the baselines and experimental settings are provided in the Appendix.

\noindent\textbf{Evaluation Metrics:} All models are evaluated using MAE and RMSE to assess urban dynamics prediction performance.

%\subsection{Evaluation Metrics}
%We evaluate urban dynamics prediction performance using MAE and RMSE. MAE is calculated as  $\mathrm{MAE} = \frac{1}{N_{r} N_{t}} \sum_{r=1}^{N_{r}} \sum_{t=1}^{N_{t}} \left| y_{r,t} - \hat{y}_{r,t} \right|$,while the RMSE is  $\mathrm{RMSE} = \sqrt{\frac{1}{N_{r} N_{t}} \sum_{r=1}^{N_{r}} \sum_{t=1}^{N_{t}} \left( y_{r,t} - \hat{y}_{r,t} \right)^{2}}$.Here, \(N_r\) denotes the number of grid cells in the region, and \(N_t\) represents the number of predicted time slots. The variables \(y_{r,t}\) and \(\hat{y}_{r,t}\) correspond to the ground-truth and predicted urban dynamics.

\subsection{Experimental Settings}
For Mask-Empowered Representation Learning and Semantic-Aware Prompting and Fine-Tuning stages, we employed the Adam optimizer with a learning rate of 0.00001 for the Shenzhen dataset and 0.0001 for the Xi’an and Chengdu datasets.
In the Test Time Adaptation stage, the Adam optimizer was applied with a learning rate of 0.00005 for Shenzhen and 0.0005 for Xi’an and Chengdu.
Additionally, LLaMA3 serves as the foundation for UrbanMind.

%\subsection{Experimental Settings}
%All experiments employ the Adam optimizer with an initial learning rate of 0.0001. For the prediction tasks, training is conducted on 50 regions, while the remaining 13 regions are reserved for testing.

\subsection{Empirical Results}
\noindent\textbf{Results of Question (1) (Zero-shot Performance)}.
% Can our UrbanMind outperform in zero-shot spatial-temporal prediction tasks for different urban dynamics?
To evaluate UrbanMind's performance in zero-shot dynamics prediction, we conducted experiments on three urban dynamics datasets across Shenzhen, Xi'an, and Chengdu. Tab.~\ref{tab:zero_shot_baselines} presents the average prediction results, while Fig.~\ref{fig:main} illustrates the detailed prediction performance over time.
UrbanMind consistently outperforms all baselines across datasets and metrics, achieving the lowest MAE and RMSE values, demonstrating outstanding generalization capability. In Shenzhen's speed prediction, while DYffusion achieved a comparable MAE to UrbanMind, its higher RMSE suggests greater prediction variance. Similarly, for Shenzhen's inflow prediction, UrbanGPT matched UrbanMind in RMSE but underperformed in MAE, highlighting UrbanMind’s superior accuracy.
UrbanMind's strength lies in its ability to generalize across diverse traffic forecasting scenarios, leveraging LLMs' generalization capability further enhanced by the test time adaptation mechanism, ensuring exceptional robustness in zero-shot urban traffic prediction tasks.
\begin{table*}[t]
\centering
\small
% \scriptsize 
\setlength{\tabcolsep}{2pt} 
\renewcommand{\arraystretch}{1.0} 
\resizebox{\textwidth}{!}{
\begin{tabular}{llcccccccccc}
\toprule
\multirow{2}{*}{\textbf{Dataset}} & \multirow{2}{*}{\textbf{Metric}} & \multirow{2}{*}{\textbf{w/o}} & \multicolumn{3}{c}{\textbf{Muffin-MAE with different masks}} & \multicolumn{2}{c}{\textbf{Spatial-Temporal Token}} & \multirow{2}{*}{\textbf{w/o}} & \multirow{2}{*}{\textbf{w/o}}& \multirow{2}{*}{\textbf{UrbanMind}} \\
\cmidrule(lr){4-6} \cmidrule(lr){7-8}
 &  & \textbf{Muffin-MAE} & \textbf{w/o Temporal} & \textbf{w/o Spatial} & \textbf{w/o Global} & \textbf{w/o Target} & \textbf{w/o Multifaceted} &  \textbf{Fine-tuning} & \textbf{Adaptation} \\
\midrule
\multirow{2}{*}{Speed} & MAE  & 0.165 & 0.144 & 0.140 & 0.141 & 0.148 & 0.153 & 0.145 & 0.142 &\textbf{0.137}\\
                       & RMSE & 0.221 & 0.168 & 0.165 & 0.168 & 0.190 & 0.189 & 0.187 & 0.166 &\textbf{0.162} \\
\multirow{2}{*}{Inflow} & MAE  & 0.250 & 0.162 & 0.175 & 0.180 & 0.189 & 0.220 & 0.143 & 0.124 & \textbf{0.114} \\
                        & RMSE & 0.343 & 0.201 & 0.205 & 0.224 & 0.247 & 0.293 & 0.209 & 0.218 &\textbf{0.173} \\
\multirow{2}{*}{Demand} & MAE  & 0.235 & 0.191 & 0.219 & 0.220 & 0.202 & 0.245 & 0.214 & 0.179 &\textbf{0.160}\\
                        & RMSE & 0.389 & 0.300 & 0.327 & 0.328 & 0.380 & 0.378 & 0.345 & 0.283 &\textbf{0.273}\\
\bottomrule
\end{tabular}
}
\caption{Ablation Study: Impacts of UrbanMind components (Muffin-MAE, LLM fine-tuning strategy, test time adaptation), masking mechanisms (temporal, spatial, global masking), and spatial-temporal token (including multifaceted and target embeddings) on zero-shot prediction accuracy. The analysis is conducted on Xi’an City.}
\label{tab:ablation_study}
\vspace{-1cm}
\end{table*}
%In the zero-shot evaluation illustrated in Fig.~\ref{fig:main}, we conducted a detailed analysis of UrbanMind's performance on 1-4 hour traffic speed predictions across three cities. Consistent with the observed overall trends, UrbanMind demonstrated superior performance, achieving the lowest RMSE values across all time horizons. These results highlight its strong capacity to generalize effectively and reliably under diverse and challenging scenarios.

\noindent\textbf{Results of Question (2) (Standard Prediction Performance)}.
% Can our UrbanMind outperform in standard spatial-temporal prediction tasks for different urban dynamics?
To evaluate UrbanMind's performance in standard spatial-temporal prediction tasks, we present the results for multiple urban dynamics across cities in Tab.~\ref{tab:same_region_baselines}. UrbanMind consistently outperforms baselines, achieving the lowest MAE and RMSE across all datasets and metrics. In Shenzhen's traffic speed prediction, UrbanMind demonstrated superior accuracy and consistency compared to DYffusion, with lower MAE and RMSE values. Similarly, in Xi'an's demand prediction, UrbanMind surpassed UrbanGPT, further highlighting its ability to model complex spatial-temporal relationships and intercorrelations among dynamics. These results establish UrbanMind as a robust and effective model for urban traffic dynamics prediction.

\noindent\textbf{Results of Question (3) (Ablation Study)}.
% (3) Is each component in UrbanMind effective when performing spatial-temporal predictions? 
We conducted an ablation study to evaluate the contribution of each component in UrbanMind, with results presented in Tab.~\ref{tab:ablation_study}. Removing Muffin-MAE leads to a significant drop in performance, with notably higher MAE and RMSE values across datasets, underscoring its importance in capturing intercorrelated spatial-temporal dependencies. Similarly, omitting LLM fine-tuning or test-time adaptation results in substantial performance degradation, highlighting their critical role in improving adaptability and precision, especially in zero-shot scenarios.
Additionally, we examine the effects of different masking mechanisms in Muffin-MAE (\eg, removing temporal, spatial, or global masking) and assess the impact of multifaceted and target embeddings in the final spatial-temporal tokens (\eg, removing either target embeddings or multifaceted embeddings) on prediction accuracy. In both cases, performance degradation is evident in increased MAE and RMSE, further validating the effectiveness of these components in UrbanMind.

\noindent\textbf{Results of Question (4) (Hyperparameters Evaluation)}.
% (4) How do different hyperparameters affect the performance? 
To analyze the impact of different hyperparameters on model performance, we conducted experiments on traffic speed prediction in Xi’an, focusing on four key hyperparameters: the temporal masking ratio \(p_t\), spatial masking ratio \(p_s\), number of trainable layers during LLM fine-tuning, and number of multifaceted dynamics. The results are presented in Fig.~\ref{fig:hyper}.  
For \(p_t\), as shown in Fig.~\ref{fig:hyper}(a), increasing the ratio to \(0.33\) achieves the lowest RMSE, while both lower and higher masking ratios degrade performance. Similarly, Fig.~\ref{fig:hyper}(b) demonstrates that the spatial masking ratio \(p_s\) achieves optimal performance at \(p_s = 0.25\), with performance gradually worsening as the ratio increases, suggesting that excessive spatial masking hinders learning.  
The effect of trainable layers during LLM fine-tuning, shown in Fig.~\ref{fig:hyper}(c), reveals that increasing the number of fine-tuned layers generally reduces RMSE, with a slight fluctuation at 4 layers, while overall, more layers help the model capture complex spatial-temporal relationships more effectively. Finally, Fig.~\ref{fig:hyper}(d) illustrates that incorporating more multifaceted dynamics consistently improves RMSE, highlighting the importance of including multifaceted embeddings in Muffin-MAE in capturing intercorrelated spatial-temporal relationships.

%% file: Content/related.tex
\section{Related Work}\label{sec:related}
\noindent \textbf{Spatial-temporal urban prediction} is essential for effective city management. Recent advancements have leveraged deep learning techniques to enhance predictive accuracy in diverse urban domains. Models such as graph neural networks~\cite{LSGCN,Yu_2018,Zheng_Fan_Wang_Qi_2020,li2017diffusion,yu2017spatio}, Transformers~\cite{li2024tsttrans,huang2024dstgtn,lin2024lvstformer,sha2024ccdsreformer,li2022sttis}, and generative adversarial networks~\cite{curbgan,ICDMtrafficGAN,zhang2021c3}, have been introduced to capture spatial-temporal patterns. Additionally, cutting-edge techniques like contrastive learning~\cite{pan2024rethinking,li2023sts,li2023urban}, and diffusion models~\cite{liu2024align,cachay2024dyffusion,zhang2023chattraffic,wang2024updiff} have further enhanced the prediction performance. Despite these innovations, most existing approaches are constrained by the need to train separate models for each specific dataset, limiting their adaptability and scalability. While some studies have explored transfer learning and meta-learning across regions~\cite{zhang2022strans,mestgan,sigkdd2022p24,ijcai2019p262,ijcai2022p282}, these methods still rely on data from the target region. In contrast, our proposed model addresses these limitations by integrating spatial-temporal data with Large Language Models, offering a generalized solution capable of adapting across diverse urban scenarios.

\noindent \textbf{Large Language Models (LLMs)} have revolutionized artificial intelligence, driving breakthroughs in text comprehension, reasoning, and generalization across diverse domains. Notable models such as ChatGPT~\cite{openai2023chatgpt}, Claude~\cite{anthropic2023claude}, LLaMA~\cite{touvron2023llama}, Vicuna~\cite{vicuna2023}, and ChatGLM~\cite{du2022glm} have gained widespread attention due to their scalability and adaptability, driving research into their diverse applications. These models have demonstrated exceptional capabilities in tasks such as graph-based reasoning~\cite{qian2023large} and multimodal learning~\cite{alayrac2022flamingo}, showcasing their potential to extend beyond traditional text-centric applications. Furthermore, their integration with advanced strategies like few-shot learning~\cite{brown2020language} has opened new avenues for addressing challenges in data-scarce domains.
Despite these advancements, the application of LLMs to zero-shot and few-shot learning in urban spatial-temporal prediction remains nascent, constrained by challenges such as adapting domain-specific data and addressing interdependencies across spatial-temporal scales.

\noindent \textbf{Spatial-Temporal Pretraining Models} have emerged as a promising approach to enhancing predictive performance in urban scenarios. Models such as UrbanGPT~\cite{UrbanGPT} and UniST~\cite{yuan2024unist} integrate spatial-temporal data with LLMs using prompt tuning. Other works, including GPT-ST~\cite{li2023gptst}, ST-LLM~\cite{ST-LLM}, TPLLM~\cite{TPLLM}, GATGPT~\cite{GATGPT}, and STG-LLM~\cite{STG-LLM}, primarily focus on graphs or grid-based worlds, adapting single spatial-temporal datasets for use with LLMs.
%%Additionally, intelligent urban systems like CityGPT~\cite{zhang2022citygpt} have demonstrated their utility in language-based tasks. 
However, these models are either ill-equipped to handle the inter-correlated multi-faceted dynamics inherent in urban environments or rely solely on the inherent generalization ability of LLMs to adapt across different scenarios, overlooking the critical issue of distributional shifts between training and testing data.

%% file: Content/appendix.tex
\section{APPENDIX FOR REPRODUCIBILITY}
To support the reproducibility of the results in this paper, we publish our code and data~\footnote{UrbanMind code: \url{https://github.com/Yliu1111/UrbanMind.git}}. Here, we describe the dataset and baseline settings in detail.

\subsection{Detailed Description of the dataset}
As previously mentioned, the dataset consists of three types of data: (1) traffic speed, (2) taxi inflow, and (3) travel demand. These data are derived from urban records collected in Shenzhen, Xi'an, and Chengdu, China. The details of the original dataset are presented in Tab.~\ref{data}. 

\noindent\textbf{Data Preprocessing for Shenzhen:} 
To expand our datasets, we applied a basic data augmentation technique by segmenting large city maps into smaller regions. Using Shenzhen as an example, the city is mapped onto a 40$\times$50 grid, which is then subdivided into smaller regions $R_{ij}$, each consisting of 10$\times$10 grid cells. Starting with region $R_{11}$, we extract the initial 10$\times$10 section. We then shift the window 5 grid cells to the right (from $s_{11}$ to $s_{16}$) to obtain the next region, $R_{12}$. By continuously sliding the window in this manner, we generate multiple overlapping regions $R_{ij}$, all maintaining the same 10$\times$10 size. The final dataset for Shenzhen has dimensions of (162, 12, 63, 10, 10), corresponding to the number of days, time slots per day, number of regions, and the width and height of each region.

\noindent\textbf{Data Preprocessing for Xi'an and Chengdu:}
To process the datasets for Xi'an and Chengdu, the urban areas are divided into 20$\times$20 grid cells, which are further partitioned into four regions $R_{ij}$ without augmentation, each consisting of 10$\times$10 grid cells. Specifically, the grid is evenly split along both horizontal and vertical axes, resulting in regions $R_{11}$, $R_{12}$, $R_{21}$, and $R_{22}$. The final dataset for both Xi'an and Chengdu has a shape of (30, 12, 4, 10, 10), representing (number of days, time slots per day, number of regions, region width, region height). We train on 3 regions and select the remaining region for testing.

\noindent\textbf{Normalization:} To address outliers in the datasets from all three cities, we implemented tailored strategies for each data type. For the traffic speed data, we imposed an upper limit of 140, with any values above this threshold clipped to 140. In the case of taxi inflow and travel demand, we defined the upper bounds based on the 90th percentile of their respective data distributions to effectively manage extreme values. Following the outlier treatment, all datasets underwent min-max normalization, scaling the data to a range between -1 and 1.

\begin{table}[h]
\vspace{-0.3cm}
%\centering
%\begin{adjustbox}{width=1\textwidth}
\large
\caption{Dataset descriptions.}
\vspace{-0.3cm}
\begin{center}
\setlength{\tabcolsep}{3.5mm}{
\scalebox{0.45}{
\resizebox{\textwidth}{!}{
\begin{tabular}{c|c|c|c}
\hline
\textbf{City}&\textbf{City size} &\textbf{Data}&\textbf{Timespan}\\
%\hline
%\textbf{Task} & traffic speed estimation &taxi inflow estimation  \\
\hline
\multirow{3}*{Shenzhen}&\multirow{3}*{$40\times50$}&Speed &07/01/16-12/31/16\\
\cline{3-4}
&&Inflow &07/01/16-12/31/16 \\
\cline{3-4}
&&Demand &07/01/16-12/31/16 \\
\hline
\multirow{3}*{Chengdu}& \multirow{3}*{$20\times20$}&Speed &10/01/16-10/31/16 \\
\cline{3-4}
&&Inflow &10/01/16-10/31/16 \\
\cline{3-4}
&&Demand &10/01/16-10/31/16 \\
\hline
\multirow{3}*{Xi'an}& \multirow{3}*{$20\times20$}&Speed &10/01/16-10/31/16 \\
\cline{3-4}
&&Inflow &10/01/16-10/31/16 \\
\cline{3-4}
&&Demand &10/01/16-10/31/16 \\
\hline
\end{tabular}}}}
\vspace{-0.5cm}
\end{center}
\label{data}
\end{table}

\noindent\textbf{Data Description:}
We evaluate our model using nine real-world urban dynamics datasets, covering three data types: (1) traffic speed, (2) taxi inflow, and (3) travel demand. Three datasets were collected in Shenzhen, China, from July 1 to December 31, 2016. The city is divided into a $40\times50$ grid, which is further aggregated into 63 regions, each consisting of $10\times10$ grid cells, covering the entire city.
The other six datasets come from Xi'an and Chengdu, China, spanning October 1 to October 31, 2016. Both cities are partitioned into $20\times20$ grids and divided into 4 regions of $10\times10$ grid cells each, where traffic speed, taxi inflow, and travel demand are measured.

Here are the details of the urban dynamics datasets:

\begin{itemize}[leftmargin=*] \item \textbf{Traffic Speed.}
In Shenzhen, the dataset includes traffic speed data for 63 regions over 4416 one-hour time slots spanning 6 months. For Xi'an and Chengdu, it covers 4 regions with 360 one-hour time slots over 1 month. Traffic speed is calculated as the ratio of travel distance to travel time for each grid cell per time slot.

\item \textbf{Taxi Inflow.}
The Shenzhen dataset records taxi inflow for 63 regions across 4416 one-hour time slots, while the Xi'an and Chengdu datasets cover 4 regions with 360 time slots each. Taxi inflow represents the number of taxi arrivals at each grid cell within a specific hour.

\item \textbf{Travel Demand.}
This dataset tracks travel demand in 63 regions of Shenzhen (4416 time slots) and 4 regions of Xi'an and Chengdu (360 time slots each). Travel demand is measured by the number of taxi pickups and drop-offs within each grid cell per hour, serving as a proxy for overall demand based on taxi data, as validated in prior studies~\cite{zhang2021c3,8970742,zhang2022strans,curbgan}. 
\end{itemize}

In short, the three datasets from Shenzhen have dimensions of $(162 \times 63, 12, 10, 10)$, where 162 represents the number of valid days, 63 is the number of regions, 12 indicates one-hour time slots per day, and $10 \times 10$ is the region size. Similarly, the six datasets from Xi’an and Chengdu are sized $(30 \times 4, 12, 10, 10)$, with 30 valid days, 4 regions, 12 one-hour time slots per day, and $10 \times 10$ grid cells per region.

\subsection{Baselines Settings}
For both the Mask-Empowered Representation Learning and Semantic-Aware Prompting and Fine-Tuning stages, we utilized the Adam optimizer with a learning rate of 0.00001 for the Shenzhen dataset and 0.0001 for the Xi’an and Chengdu datasets. During the Test Time Adaptation stage, the Adam optimizer was also employed, with learning rates set to 0.00005 for Shenzhen and 0.0005 for Xi’an and Chengdu. Moreover, LLaMA3 served as the backbone model for UrbanMind. The detailed architectures and configurations of the baseline models are outlined below.
\begin{itemize}
\item\textbf{GATGPT\cite{chen2023gatgpt}}:
GATGPT integrates a pre-trained GPT-2 model with a graph attention mechanism to capture complex spatial-temporal dependencies for traffic prediction. The model first applies a graph attention layer to extract spatial relationships by learning adaptive weights based on region adjacency matrices. The output is then projected and fed into the GPT-2 model to capture temporal dynamics. Finally, a linear decoding layer maps the GPT-2 outputs to the target spatial grid.

\item\textbf{GCNGPT\cite{chen2023gatgpt,liu2024spatial}}: 
GCNGPT integrates graph convolutional networks (GCNs) with a pre-trained GPT-2 model to effectively capture spatial-temporal dependencies in traffic data. The model first applies a GCN to extract spatial features from region-wise adjacency matrices, transforming the input into a representation that encodes spatial relationships. This output is then reshaped and fed into the GPT-2 model to model complex temporal dynamics. A linear projection layer maps the GPT-2 outputs to the target grid for final predictions.

\item \textbf{STLLM\cite{liu2024spatial}}:
ST-LLM uses large language models for traffic prediction by combining spatial-temporal embeddings with a partially frozen attention strategy to capture global dependencies. The baseline integrates spatial and temporal features to enhance traffic dynamics representation, leveraging LLMs to model complex patterns. Key attention layers are partially frozen to ensure stability while allowing selective fine-tuning for traffic-specific tasks. The model processes inputs with embedded time features, and predictions are made through a transformer-based architecture optimized with mean squared error loss.

\item \textbf{TPLLM\cite{ren2024tpllm}}: 
TP-LLM integrates temporal and spatial information for traffic prediction using a hybrid approach that combines sequence embeddings, graph embeddings, and a GPT-2 backbone. The model starts by transforming the input data through a sequence embedding layer using 1D convolutions and a graph embedding layer via linear transformations to capture temporal patterns and spatial relationships, respectively. These embeddings are then fused and passed through a GPT-2 model configured with six layers and eight attention heads, enabling the extraction of complex temporal dependencies. The output from GPT-2 is processed through a linear layer to map it to the desired prediction dimensions.

\item \textbf{UrbanGPT\cite{li2024urbangpt}}:
UrbanGPT integrates a spatio-temporal dependency encoder with instruction-tuning to enable large language models (LLMs) to handle complex time-space interdependencies and generalize across diverse urban tasks. The baseline model employs the spatial-temporal encoder (ST\_Enc) in conjunction with LLaMA2\cite{touvron2023llama} to capture dynamic patterns in urban data, effectively modeling both spatial relationships and temporal trends. The ST\_Enc module incorporates dilated inception convolution layers to extract multi-scale spatiotemporal features, which are then processed through LLaMA2 to capture global dependencies. Instruction-tuning enhances the model's adaptability.

\item \textbf{STG-LLM\cite{liu2024can}}: 
STG-LLM integrates a spatial-temporal graph tokenizer with a lightweight adapter to process complex spatial-temporal data effectively. The model begins with a linear encoder that transforms raw spatial-temporal inputs into a latent representation. Temporal and weekly patterns are captured through time-of-day and day-of-week embedding layers, which provide temporal positional context. These encoded features are then combined with positional embeddings and passed through a Transformer encoder, simulating an GPT-2 to model long-range dependencies across time and space. Finally, a linear decoder refines the output.

\item \textbf{LLaMA3\cite{dubey2024llama}}:
LLaMA3 is a foundation model designed for multilingual understanding and reasoning. Its advanced capabilities make it ideal for traffic dynamics prediction. The baseline model leverages LLaMA3 to capture complex temporal dependencies in spatial-temporal data. The model architecture integrates LLaMA3 as a feature extractor, with its parameters frozen except for select layers to maintain efficiency. The extracted features are processed through a linear layer to predict future traffic conditions, reshaping the output to match the spatial grid structure. Specifically, we utilized LLaMA 3.2, a foundational large language model developed by Meta, as the baseline model for UrbanMind.

\item\textbf{DYffusion\cite{cachay2024dyffusion}} 
DYffusion is a diffusion model designed for probabilistic spatial-temporal forecasting, aiming to generate stable and accurate rollout predictions. The baseline integrates a Unet architecture with a Gaussian Diffusion process to improve performance. The Unet processes single-channel inputs through an initial convolutional layer and uses multiple convolutional layers with scaled dimensions. The Gaussian Diffusion module refines predictions over time through iterative denoising steps.

\item \textbf{TGC-LSTM\cite{cui2019traffic}}: 
TGC-LSTM addresses spatial-temporal forecasting in traffic networks by modeling time-varying patterns and complex spatial dependencies. It combines traffic graph convolution and spectral graph convolution within an LSTM framework to capture both spatial and temporal dynamics. The LSTM models temporal dependencies, while convolutional layers extract spatial features from the traffic network. The model flattens spatial dimensions for LSTM processing and reconstructs outputs to retain spatial structure. It is trained using mean squared error loss.

\item \textbf{GCRN\cite{seo2018structured}}: 
GCRN extends classical RNNs to handle arbitrary graph-structured data, enabling effective modeling of spatial and temporal dependencies. The baseline model integrates a graph convolutional layer with a gated recurrent unit (GRU) to capture both spatial relationships among nodes and temporal dynamics over time. The graph convolutional layer processes input features to learn spatial dependencies based on the adjacency matrix, while the GRU captures temporal patterns from sequential data. The final output is generated through a fully connected layer that predicts future traffic conditions.

\item \textbf{GAGCN\cite{xia2022short}}: 
GAGCN uses graph attention networks to extract and dynamically adjust spatial associations among nodes over time. The baseline builds on this by integrating graph attention into a convolutional neural network to capture spatial-temporal dependencies. Input traffic data is processed through stacked convolutional layers to learn local spatial features, followed by graph attention layers that refine node relationships based on temporal patterns. Final predictions are made through fully connected layers, and the model is optimized with mean squared error loss.

\end{itemize}

\section{Appendix B: Additional Experiments}
\label{app:efficiency}
\subsection{Computational Efficiency Details}

We report the runtime cost of our LLaMA3-based experiments on the Shenzhen dataset. UrbanMind takes 70.9 seconds per epoch for training and 16.5 seconds per epoch for test-time adaptation. In comparison, the baseline UrbanGPT (also based on LLaMA) requires 80.1 seconds per epoch for training. Although UrbanMind introduces moderate additional cost from Muffin-MAE and the adaptation module, it remains practical and suitable for real-time inference scenarios, especially considering its improved generalization capability.

\subsection{Cross-City Generalization}

To evaluate cross-city generalization, we conduct a zero-shot transfer experiment where models are trained on traffic speed data from Shenzhen and directly evaluated in Xi’an without fine-tuning. We specifically compare UrbanMind against UrbanGPT. UrbanMind achieves 8.5\% lower MAE (0.194 vs. 0.212) and 9.9\% lower RMSE (0.236 vs. 0.262), demonstrating its superior adaptability to unseen urban environments.

\subsection{Additional Baseline}
We include STGAIL~\cite{liu2024align} as a new baseline on traffic speed prediction in Xi’an. STGAIL achieves an MAE of 0.227 and RMSE of 0.266, underperforming compared to UrbanMind (0.194 / 0.236). This suggests its limited capacity for handling multimodal spatial-temporal signals in urban forecasting.